\documentclass{article}

\PassOptionsToPackage{numbers, compress}{natbib}

\usepackage[preprint]{neurips_2023}

\usepackage[utf8]{inputenc} 
\usepackage[T1]{fontenc}    
\usepackage{hyperref}       
\usepackage{url}            
\usepackage{booktabs}       
\usepackage{amsfonts}       
\usepackage{nicefrac}       
\usepackage[misc]{ifsym}
\usepackage{microtype}      
\usepackage{xcolor}         
\usepackage{enumitem}
\setlist{nolistsep, leftmargin=0.1in}

\usepackage{graphicx}
\usepackage{amsmath}
\usepackage{amssymb}
\usepackage{booktabs}
\usepackage{multirow}
\usepackage{makecell}
\usepackage{floatrow}
\usepackage{caption}
\usepackage{tabularx}
\usepackage{xltabular}
\usepackage{longtable}
\usepackage{subcaption}

\usepackage{framed}
\newbox\totalbox
\newbox\partialbox
\newdimen\partialboxdim

\usepackage{chngcntr}

\newenvironment{continueframe}{%
    \advance\linewidth-2\fboxsep
    \advance\linewidth-2\fboxrule
    \hsize=\linewidth
    \partialboxdim=\dimexpr\pagegoal-\pagetotal-\pageshrink-6pt-\baselineskip\relax
    \setbox\totalbox=\vbox\bgroup\begingroup
}{%
    \endgraf\endgroup\egroup
    \setbox\partialbox=\vsplit\totalbox to\partialboxdim
    \par\smallskip
    \hbox{\fbox{\vbox{\unvbox\partialbox}}}\nopagebreak
    
}
\title{LVLM-eHub: A Comprehensive Evaluation Benchmark for Large Vision-Language Models}

\author{Peng Xu$^{*\dagger2, 1}$, Wenqi Shao$^{*1}$, Kaipeng Zhang$^{*1}$, Peng Gao$^{*1}$, Shuo Liu$^{1}$, \\ \textbf{Meng Lei}$^{\dagger1,3}$\textbf{,} \textbf{Fanqing Meng}$^{\dagger1}$\textbf{,}  \textbf{Siyuan Huang}$^{\dagger1}$\textbf{,}  \textbf{Yu Qiao}$^{1}$\textsuperscript{\Letter}\textbf{,} \textbf{Ping Luo}$^{2, 1}$\textsuperscript{\Letter}\\\\
$^{1}$OpenGVLab, Shanghai AI Laboratory  $^2$The University of Hong Kong  \\
$^{3}$Peking University
}

\begin{document}

\maketitle

\renewcommand{\thefootnote}{\fnsymbol{footnote}}
{\let\thefootnote\relax\footnotetext{
\noindent \hspace{-5mm}$^*$ Equal First Authors\\
\textsuperscript{\Letter} Corresponding Authors: qiaoyu@pjlab.org.cn; pluo@cs.hku.edu \\
$^\dagger$  This work was done during the internship at Shanghai AI Laboratory. }}

\begin{abstract}
Large Vision-Language Models (LVLMs) have recently played a dominant role in multimodal vision-language learning. Despite the great success, it lacks a holistic evaluation of their efficacy. This paper presents a comprehensive evaluation of publicly available large multimodal models by building a LVLM evaluation Hub (LVLM-eHub). Our LVLM-eHub consists of $8$ representative LVLMs such as InstructBLIP and MiniGPT-4, which are thoroughly evaluated by a quantitative capability evaluation and an online arena platform. The former evaluates $6$ categories of multimodal capabilities of LVLMs such as visual question answering and embodied artificial intelligence on $47$ standard text-related visual benchmarks, while the latter provides the user-level evaluation of LVLMs in an open-world question-answering scenario. The study reveals several innovative findings. First, instruction-tuned LVLM with massive in-domain data such as InstructBLIP heavily overfits many existing tasks, generalizing poorly in the open-world scenario. Second, instruction-tuned LVLM with moderate instruction-following data may result in object hallucination issues (i.e., generate objects that are inconsistent with target images in the descriptions). It either makes the current evaluation metric such as CIDEr for image captioning ineffective or generates wrong answers. Third, employing a multi-turn reasoning evaluation framework can mitigate the issue of object hallucination, shedding light on developing an effective pipeline for LVLM evaluation. The findings provide a foundational framework for the conception and assessment of innovative strategies aimed at enhancing zero-shot multimodal techniques. Our LVLM-eHub will be available at \href{https://github.com/OpenGVLab/Multi-Modality-Arena/tree/main/LVLM_evaluation}{vlarena page}.


\end{abstract}

\section{Introduction}
Large Language Models (LLMs), such as LLaMA~\cite{touvron2023llama}, GPT-3~\cite{brown2020language}, and Vicuna~\cite{vicuna}, have demonstrated remarkable progress in Natural Language Processing (NLP). These models leverage large-scale pre-training data and huge networks to achieve impressive results in NLP benchmarks. Recently, GPT-4 \cite{gpt4} further expanded the impact to the multimodal community, stimulating the rapid development of Large Vision-Language Models (LVLMs) and revolutionizing the landscape of artificial intelligence.


Large Vision-Language Models (LVLM) have achieved remarkable progress in multimodal vision-language learning for various multimodal tasks such as visual question answering and multimodal conversation. Specifically, LVLMs capitalize on the knowledge from LLMs and effectively align visual features with the textual space. Flamingo~\cite{alayrac2022flamingo}, a pioneering LVLM, integrates visual features into LLMs through cross-attention layers. Later studies proposed more efficient vision-text interactions~\cite{li2023blip}, more efficient training methods~\cite{gao2023llama,zhang2023vpgtrans}, and employing instruction tuning~\cite{liu2023visual,gao2023llama,zhu2023minigpt,ye2023mplug,li2023otter,dai2023instructblip,zhang2023vpgtrans}. 

However, despite the great success, few efforts have been made to provide systematic evaluations of LVLMs. But evaluation plays a critical role in understanding the strengths and weaknesses of LVLMs, thereby guiding their future development. Recent work \cite{li2023evaluating} presents a systematic investigation of object hallucination of LVLMs by proposing a polling-based object probing evaluation method. Moreover, ImageNetVC \cite{xia2023imagenetvc} studies how well LVLMs can master visual commonsense knowledge. Liu et al. \cite{liu2023hidden} comprehensively evaluate the performance of LVLMs in visual recognition with text recognition, such as optical character recognition. GVT \cite{wang2023makes} evaluates
LVLM’s visual semantic understanding and fine-grained perception capabilities. Nevertheless, these studies only evaluate a portion of LVLMs on specific tasks, lacking an overall understanding of LVLM's capabilities.

\begin{figure*}[t]
		\centering
		\includegraphics[width=0.9\linewidth]{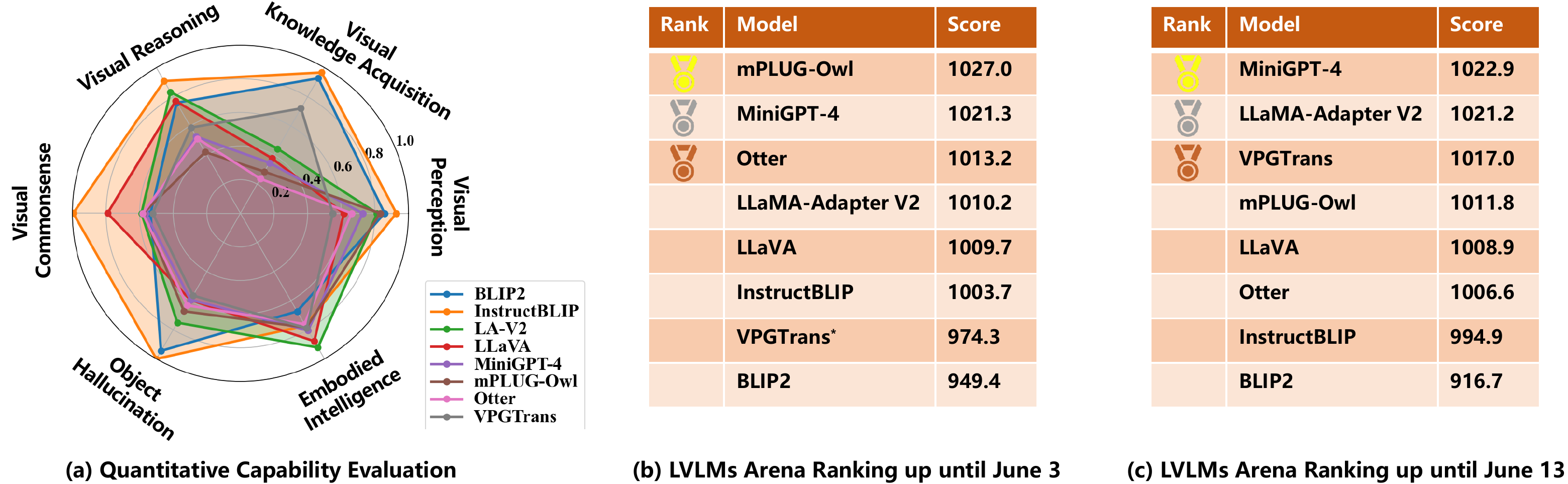}
		\caption{Comparison of LVLMs in the LVLM-eHub. (a) illustrates the variances in quantitative capability performance across six distinct aspects among LVLMs (see ranking list in Fig. \ref{fig:lvlm-radar-list}). (b) and (c) present the Elo rating ranking of LVLMs in the LVLM Arena according to the data collected up until June 3 and June 13, respectively. The arena ranking list changes as more evaluation samples are involved. We see that InstructBLIP performs well in existing in-domain tasks, while instruction-tuned models with moderate high-quality instruction-following data such as mPLUG-Owl, MiniGPT-4, Otter and LLaMA-Adapter V2 work well in the open-world arena. VPGTrans$^*$ means that VPGTrans adopts the prompt in the same form with MiniGPT-4.}
		\label{fig:lvlm-radar}
\end{figure*}

In pursuit of a comprehensive evaluation of LVLMs, we build an LVLM Evaluation hub (LVLM-eHub) consolidating $8$ representative LVLMs such as InstrucBLIP \cite{dai2023instructblip} and MiniGPT-4 \cite{zhu2023minigpt}. The details about model configuration and training data are listed in Table \ref{tab:comparison-LVLMs}. 
Our LVLM-eHub consists of a quantitative capability evaluation and an online arena platform, providing a thorough investigation of the selected LVLMs. 
Specifically, the quantitative capability evaluation extensively evaluates 6 categories of multimodal capabilities of LVLMs including visual perception, visual knowledge acquisition, visual reasoning, visual commonsense, object hallucination, and embodied intelligence (see Fig. \ref{fig:lvlm-radar} (a)), by collecting $47$ standard text-related visual benchmarks. On the other hand, the online arena platform features anonymous randomized pairwise battles in a crowd-sourced manner, providing a user-level model ranking in the open-world question-answering scenario (see Fig. \ref{fig:lvlm-radar} (b \& c)).



Our LVLM-eHub comprehensively evaluates LVLMs, revealing several innovative findings. (1) Instruction-tuned LVLM with massive in-domain data suffers from overfitting problem and generalizes poorly in open-world scenarios , such as InstructBLIP (see Fig. \ref{fig:lvlm-radar} (a)). (2) With moderate instruction-following data, instruction-tuned LVLM may cause object hallucination issues, generating objects that are inconsistent with target images in the descriptions. This leads to incorrect answers or renders current evaluation metrics, such as CIDEr for image captioning, ineffective. (3) We find that a multi-turn reasoning evaluation pipeline can mitigate the issue of object hallucination, indicating that developing an effective pipeline for LVLM evaluation is urgent. 

The contributions of our work are summarized follows. (1) We propose LVLM-eHub which is the first comprehensive evaluation benchmark for large vision-language models, to our best knowledge. (2) LVLM-eHub provides extensive evaluation on 6 categories of multimodal capabilities of LVLMs in $47$ text-based visual tasks. (3) LVLM-eHub builds an online arena platform for LVLMs, which features anonymous randomized pairwise user-level comparison in a open-world scenario. (4) Our evaluation results reveal several innovative findings, providing a
foundational framework for the assessment of innovative strategies aimed at enhancing zero-shot multimodal techniques.

\begin{table*}[t]
\centering
\small
\caption{\textbf{Comparison between Different LVLMs.} `VE', `Adapt', `ToP', `TuP', and `\# Token' represent the visual encoder, adaption module, number of total parameters, tuning parameters, and visual tokens fed into the LLM, respectively. 
$^\dag$ indicates that the model is frozen. CC$^*$ consists of COCO \cite{chen2015microsoft}, CC3M \cite{sharma2018conceptual}, and CC12M \cite{changpinyo2021conceptual}.  CC, VG, CY, L400, and LC indicate Conceptual Caption \cite{sharma2018conceptual,changpinyo2021conceptual}, Visual Genome~\cite{krishna2017visual}, COYO-700M \cite{kakaobrain2022coyo-700m}, LAION 400M~\cite{schuhmann2021laion} and LAION COCO~\cite{laion_coco}, respectively. SBU \cite{ordonez2011im2text} contains 1 million images with captions. LLaVA-I represents 158K multimodal instruction-following data in LLaVA \cite{liu2023visual}. QA$^*$ denotes 13 question-answering datasets in InstructBLIP \cite{dai2023instructblip}. We count all the data and tuning parameters needed to convert the pretrained vision model and LLM into a visual instruction model. }
\label{tab:comparison-LVLMs}\vspace{-0.2in}
\resizebox{\textwidth}{!}{%
\begin{tabular}{c|cccccc|cc|cc}
\toprule
\multirow{2}{*}{Model} & \multicolumn{6}{c|}{Model Configuration} & \multicolumn{2}{c|}{Image-Text Data} & \multicolumn{2}{c}{Visual Instruction Data}  \\
\cmidrule{2-11}
& VE      & LLM    & Adapt &ToP &TuP &\# Token      & Source  & Size     & Source                  & Size      \\
\cmidrule{1-11}
BLIP2 & ViT-g/14$^\dag$      & FlanT5-XL$^\dag$   & Q-Former &4B &107M & 32      & CC$^*$-VG-SBU-L400  & 129M     & -                  & - \\
LLaVA & ViT-L/14$^\dag$     & Vicuna   & FC layer &7B &7B & 256     & CC3M  & 595K     & LLaVA-I                  & 158K \\
LA-V2 & ViT-L/14$^\dag$      & LLaMA$^\dag$   & B-Tuning &7B &63.1M &10      & L400  & 200M     & LLaVA-I                 & 158K \\
MiniGPT-4 & BLIP2-VE$^\dag$   & Vicuna$^\dag$   & FC layer &7B & 3.1M & 32      & CC-SBU-L400  & 5M     & CC+ChatGPT                 & 3.5K \\
mPLUG-Owl &ViT-L/14 & LLaMA$^\dag$ & LoRA &7B &388M & 65      & CC$^*$-CY-L400  & 204M    & LLaVA-I                 & 158K \\ 
Otter &ViT-L/14$^\dag$ & LLaMA$^\dag$ & Resampler &9B & 1.3B & 64      & -  & -    & LLaVA-I                & 158K \\ 
InstructBLIP &ViT-g/14$^\dag$ & Vicuna$^\dag$ & Q-Former &7B & 107M & 32      & -  & -    & QA$^*$                & 16M \\ 
VPGTrans &ViT-g/14$^\dag$ & Vicuna$^\dag$ & Q-Former &7B & 107M & 32      & COCO-VG-SBU-LC  & 13.8M    &CC+ChatGPT                 & 3.5K \\ 
\bottomrule
\end{tabular}%
}
\end{table*}

\section{LVLM Evaluation Hub}

\begin{figure*}[t]
		\centering
		\includegraphics[width=0.9\linewidth]{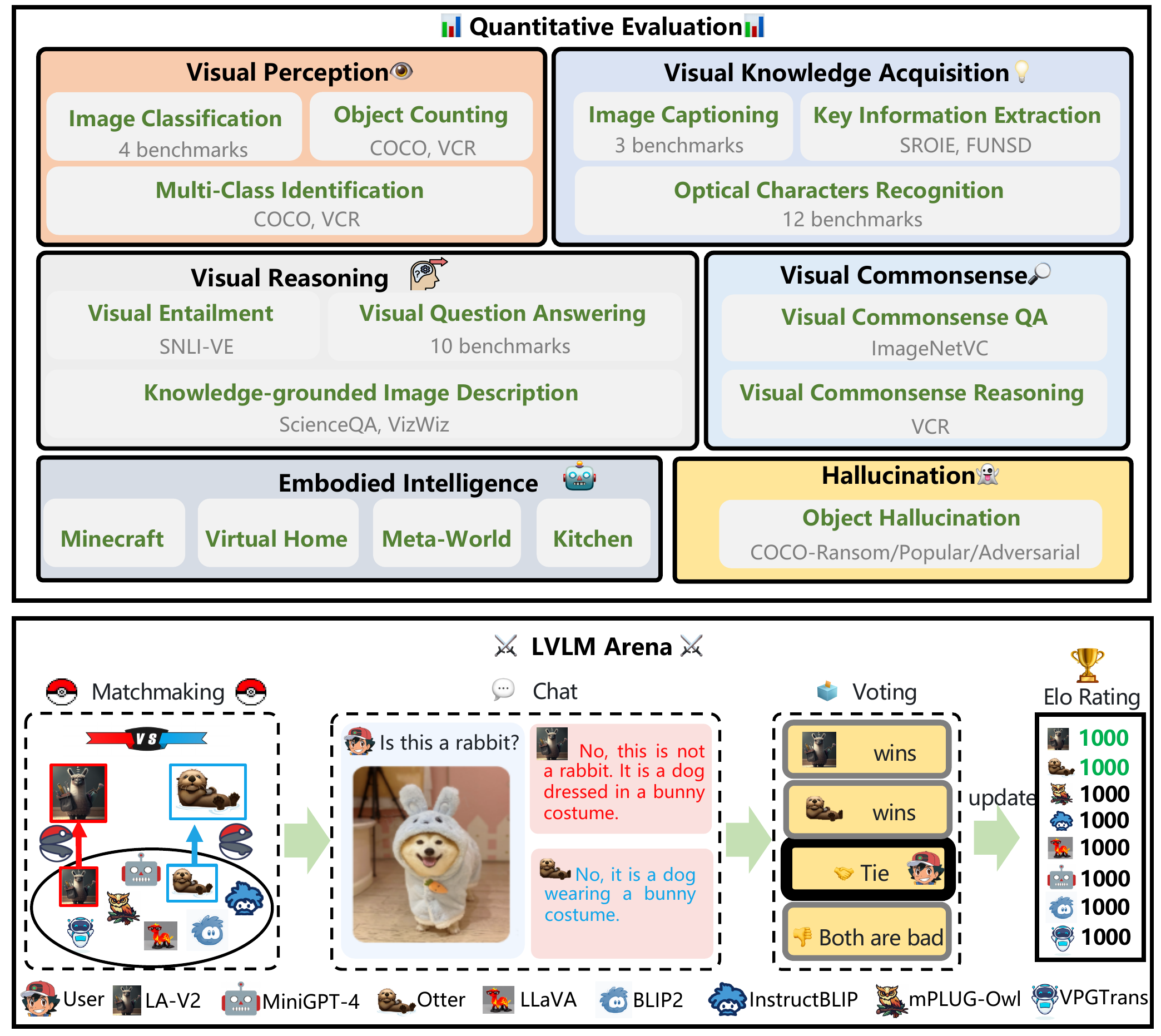}
		\caption{Our evaluation encompasses quantitative evaluation and online LVLM Arena. Plentiful benchmarks are employed to comprehensively evaluate the six critical capabilities of the models in the quantitative evaluation. In the LVLM Arena, an online platform, users can participate in an online evaluation by chatting with two anonymous models and choosing their preferred model.}
		\label{fig1}
\end{figure*}

In this section, we introduce representative LVLMs, multimodal capabilities of interest, and evaluation methods. The whole LVLM Evaluation Hub is illustrated in Fig. \ref{fig1}. Our LVLM evaluation hub compromises $8$ representative models
including BLIP2 \cite{li2023blip}, LLaVa \cite{liu2023visual}, LLaMA-Adapter V2 \cite{gao2023llama}, MiniGPT-4 \cite{zhu2023minigpt}, mPLUG-Owl \cite{ye2023mplug}, Otter \cite{li2023otter}, InstructBLIP \cite{dai2023instructblip}, and VPGTrans \cite{zhang2023vpgtrans}. 
All models boost vision-language representation learning by utilizing pre-trained image encoders and large language models (LLM). But they differ in training data scale and model configuration as shown in Table \ref{tab:comparison-LVLMs}, where the information is collected from their papers or provided by the authors. For a fair comparison between LVLMs, we collect their checkpoints with parameter sizes less than $10$B. The detailed descriptions of these models are provided in Appendix Sec. \ref{appendix:lvm_model_hub}.

\subsection{Quantitative Capability Evaluation}\label{sec:method-capability}

We aim to evaluate LVLMs' capability comprehensively. In particular, we summarize $6$ categories of capabilities and collect corresponding benchmarks for quantitative evaluation (see Fig. \ref{fig1}). Please see our Appendix Sec. \ref{appendix:dataset} for more statistics and details of the collected benchmarks.

\textbf{Visual Perception.} Visual perception is the ability to recognize the scene or objects in images, the preliminary ability of the human visual system. We evaluate this capability of models through image classification (ImgCLs) using the ImageNet1K~\cite{imagenet}, CIFAR10~\cite{Krizhevsky2009LearningML}, Pets37~\cite{parkhi12a} and Flowers102~\cite{Nilsback08} benchmarks, multi-class identification (MCI) and object counting (OC) using the GVT \cite{gvt} benchmark. ImgCLs and MCI measure how well an LVLM grasps high-level semantic information, while OC assesses the recognition ability for fine-grained objects.

\textbf{Visual Knowledge Acquisition.} Visual knowledge acquisition entails understanding images beyond perception to acquire knowledge. This evaluation is conducted through Optical Characters Recognition (OCR) using twelve benchmarks (including IIIT5K~\cite{ittt5k}, IC13~\cite{ic13}, IC15~\cite{ic15}, Total-Text~\cite{total-text}, CUTE80~\cite{cute80}, SVT~\cite{svt}, SVTP~\cite{svtp}, COCO-Text~\cite{coco-text}, WordArt~\cite{wordart}, CTW~\cite{ctw}, HOST~\cite{ost}, WOST~\cite{ost}), Key Information Extraction (KIE) using the SROIE~\cite{sroie} and FUNSD~\cite{funsd}, and Image Captioning (ImgCap) using two benchmarks (including NoCaps~\cite{nocaps} and Flickr30K~\cite{flickr30k}). The OCR task measures whether a model can accurately identify and extract text from images or scanned documents. The KIE task further poses challenges in extracting structured information from unstructured or semi-structured text. Finally, ImgCap assesses whether a model can generate a good natural language description of the content of an image.


\textbf{Visual Reasoning.} Visual reasoning requires a comprehensive understanding of images and related texts. To evaluate the visual reasoning ability of LVLMs, we utilize three tasks including visual question answering (VQA) (including DocVQA\cite{docvqa}, TextVQA\cite{textvqa}, STVQA\cite{stvqa}, OCR-VQA\cite{ocr-vqa}, OKVQA\cite{okvqa}, GQA\cite{gqa}, IconQA\cite{iconqa}, Visual Spatial Reasoning (VSR)\cite{vsr}, and Visual Dialog (Visdial)\cite{visdial}.), knowledge-grounded image description (KGID), and visual entailment. For KGID, we use ScienceQA~\cite{scienceqa} and VizWiz~\cite{vizwiz}). For visual entailment task, we use SNLI-VE. These three tasks are in VQA form. A capable LVLM should be able to understand the objects and scenes in an image and can reason to generate answers that are semantically meaningful to the question asked.


\textbf{Visual Commonsense.} Visual commonsense refers to the general visual knowledge commonly shared across the world, as opposed to the visual information specific to a single image. This evaluation tests the model’s understanding of commonly shared human knowledge about generic visual concepts using ImageNetVC~\cite{xia2023imagenetvc} and visual commonsense reasoning (VCR)~\cite{zellers2019recognition}. Specifically, ImageNetVC is utilized for zero-shot visual commonsense evaluation, such as color and shape, while VCR covers various scenes, such as spatial, casual, and mental commonsense.


\textbf{Object Hallucination.} LVLM  suffers from the object hallucination problem, i.e., the generated results are inconsistent with the target images in the descriptions \cite{li2023evaluating}. Evaluating object hallucination for different LVLMs help understand their respective weaknesses. To this end, we evaluate the object hallucination problem of LVLMs on the MSCOCO dataset \cite{chen2015microsoft} under POPE pipeline \cite{li2023evaluating}.

\textbf{Embodied Intelligence.} Embodied intelligence aims to create agents, such as robots, which learn to solve challenging tasks requiring environmental interaction. Recently, LLM and LVLM exhibited exceptional effectiveness in guiding the agent to complete a series of tasks. In this evaluation, we utilize high-level tasks in EmbodiedGPT~\cite{embodiedgpt} and employ Minecraft~\cite{fan2022minedojo}, VirtualHome~\cite{puig2018virtualhome}, Meta-World~\cite{gupta2019relay}, and Franks Kitchen~\cite{gupta2019relay} as benchmarks.



\subsection{Online Evaluation with LVLM Arena}
Designing quantitative evaluations for LVLM to satisfy all capabilities is challenging, as evaluating LVLM responses constitutes an open-ended problem.
Inspired by FastChat~\cite{zheng2023fastchat}, we introduce the LVLM Arena, an online evaluation framework for LVLMs' pairwise battle with human judgment.  

Figure~\ref{fig1} illustrates the LVLM Arena, comprising three primary components: matchmaking, chat, and voting. Initially, two models are sampled from the model zoo. Users then converse side-by-side with the models, who remain anonymous. Subsequently, users vote for the superior model.

\noindent\textbf{Matchmaking.} The matchmaking module samples two models in a tournament style based on their Elo rating. However, due to the currently limited size of the model hub, we employ random sampling.

\noindent\textbf{Chat.} Users chat side-by-side with two sampled models (which remain anonymous) using images or text inputs. Different from quantitative evaluation, users can chat about anything. Our existing online platform supports only single-round chats due to high computational and memory demands in multi-round chats. Future updates will address this constraint.

\noindent\textbf{Voting.} After the chat session, users vote for their preferred model. Four options are available: Model A, Model B, Tie, and Both are bad. The Elo rating is subsequently updated using voting results.

In contrast to limited quantitative evaluations, the LVLM Arena provides an open-world evaluation framework that enables users to chat with models about anything, emulating real-world conditions. Besides, users serve as the judge for the battle, which brings more convincing evaluation results than traditional evaluation metrics. 

\subsection{Zero-shot Evaluation}\label{sec:evaluation methods}
\begin{figure*}[t]
		\centering
		\includegraphics[width=0.9\linewidth]{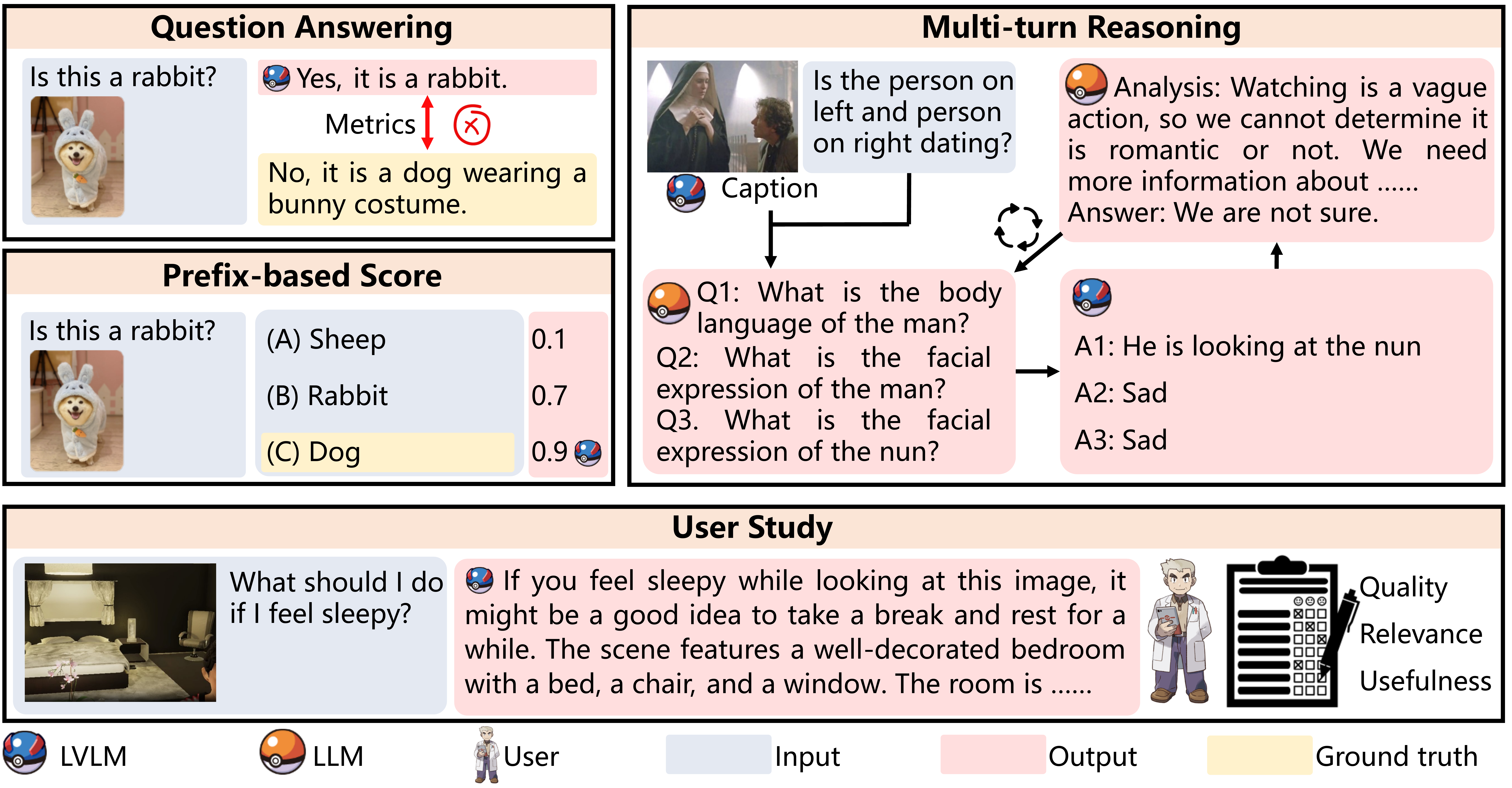}
		\caption{Illustration of our adopted evaluation methods. To evaluate the zero-shot performance of LVLMs on diverse downstream tasks,
  we employ four methods including question answering, prefix-based score, multi-turn reasoning, and user study.}
		\label{fig:evalation-methods}
\end{figure*}

LVLMs are capable of capturing a wide range of multimodal patterns and relationships. We evaluate the aforementioned $6$ categories of capabilities of LVLMs in Sec. \ref{sec:method-capability} by investigating their zero-shot performance on various  tasks. Zero-shot evaluation allows us to evaluate the LVLMs' ability to generalize to new tasks without training the model, which is competent for large-scale evaluation. To be specific, we treat the zero-shot evaluation as various forms of prompt engineering for different tasks (see Fig. \ref{fig:evalation-methods}) as presented in the following.

\begin{itemize}

\item \textit{Question Answering.} Prompting with visual question answering can be used to solve many downstream tasks, which assess how well an LVLM understands the underlying language and visual features. We design proper prompts to ensure that the LLM can produce meaningful results. For example, text prompts of OCR can be "\textit{what is written in the image?}". Then, we evaluate the answers generated by the LLM using the corresponding metric such as accuracy.

\item \textit{Prefix-based Score.} For multi-choice QA tasks, we can utilize a visual encoder to obtain visual prompts for a given image. Then, the visual prompts are prefixed into the text embeddings, which are fed into the LLM. The likelihood of image-text pair can be generated, which is referred to as a prefix-based score. We can obtain a prefix-based score for each text prompt of the candidate's answer. The answer with the largest prefix-based score is selected as the final answer.

\item \textit{Multi-turn Reasoning.} Following IdealGPT~\cite{liu2023hidden}, we use a multi-turn reasoning framework to evaluate complex visual reasoning tasks. Specifically, we utilize an LLM such as ChatGPT to generate sub-questions for a given question, an LVLM to provide corresponding sub-answers, and another LLM to evaluate sub-answers' quality. Such a pipeline iteratively proceeds until a satisfactory answer is obtained.  

\item \textit{User Study.} 
Evaluating the quality of the text generated by an LVLM  requires a thorough understanding of the underlying language and context. In embodied artificial intelligence tasks, the LVLM generates a plan for the given instruction, which should be evaluated through various aspects such as recognition accuracy and conciseness in answers. It is hard to implement such an evaluation using an existing metric. Thus, user studies are conducted to assess the quality, relevance, and usefulness of the text generated by the LVLM in a specific context. To maintain evaluation fairness, we randomly shuffle the model's output order and anonymize outputs during evaluation.
     
\end{itemize}

\begin{table*}[t]
\centering
\small
\caption{Evaluation results of visual perception capability of LVLMs on tasks of Image Classification (Imgcls), Object Counting (OC), and Multi-class Identification (MCI). We use accuracy metric for all tasks. The \textbf{best} result is \textbf{bold} while the \underline{second} is \underline{underlined}. The average score is obtained by normalizing over each row and taking the average of each column. S-SOTA indicates the supervised state-of-the-art results.}
\label{tab:oc_mci_results}
\resizebox{\textwidth}{!}{%
\begin{tabular}{c|c|ccccccccc}
\toprule
\multicolumn{2}{c|}{Datasets} & BLIP2 & InstructBLIP & LA-V2 & LLaVA & MiniGPT-4 & mPLUG-Owl & Otter & VPGTrans & S-SOTA \\
\cmidrule(lr){1-2}\cmidrule(lr){3-11}
\multirow{3}{*}{ImgCls}
& ImageNet1K & 23.71 & 24.51 & \underline{25.89} & 23.50 & 21.17 & \textbf{26.68} & 19.29 & 19.75 & 91.10~\cite{chen2023symbolic} \\ 
& CIFAR10 & 58.20 & \underline{67.24} & 64.86 & \textbf{67.96} & 61.39 & 59.66 & 65.42 & 60.88 & 99.70~\cite{kabir2023reduction} \\ 
& Pets37 & 34.79 & \underline{38.86} & 24.39 & 9.09 & 18.90 & \textbf{43.16} & 5.79 & 10.88 & 96.70~\cite{oquab2023dinov2} \\ 
& Flowers102 & 19.44 & 21.78 & \underline{22.34} & 8.38 & 21.70 & \textbf{22.91} & 6.13 & 7.97 & 99.64~\cite{IELT10042971} \\
\cmidrule(lr){1-2}\cmidrule(lr){3-11}
\multirow{2}{*}{OC}
& COCO  & \textbf{48.90} & \underline{46.65} & 38.50 & 20.56 & 20.86 & 34.14 & 46.14 & 27.30 & - \\
& VCR  & 25.05 & \underline{29.29} & 26.51 & 24.60 & 25.26 & 27.98 & \textbf{41.06} & 19.46 & -\\
\cmidrule(lr){1-2}\cmidrule(lr){3-11}
\multirow{2}{*}{MCI}
& COCO  & \underline{86.06} & \textbf{87.81} & 82.90 & 49.66 & 72.70 & 58.30 & 51.03 & 52.34 & - \\
& VCR   & 66.59 & \textbf
{76.49} & 50.66 & \underline{66.90} & 66.02 & 55.56 & 51.60 & 48.80 & - \\
\cmidrule(lr){1-2}\cmidrule(lr){3-11}
\multicolumn{2}{c|}{Average Score}
& \underline{0.858} & \textbf{0.928} & 0.813 & 0.615 & 0.727 & 0.831 & 0.661 & 0.563 & - \\
\bottomrule
\end{tabular}%
}
\end{table*}

\section{Experiment and Analysis}\label{sec:exp}
In this section, we perform a zero-shot evaluation to assess the $6$ categories of capabilities of LVLMs. Specifically, visual perception ability, visual knowledge acquisition, visual reasoning, visual commonsense understanding,  visual object hallucination, and embodied intelligence are assessed in Sec. \ref{sec:exp-perception} $\sim$ Sec.\ref{sec:exp-embodied}, respectively. The LVLM arena evaluation result is presented in Sec.\ref{sec:exp-arena}. More evaluation findings, evaluation details, quantitative results, and details about evaluation datasets can be found in Appendix Sec. \ref{appendix:lvm_model_hub}, Sec. \ref{appendix:exp}, Sec. \ref{appendix:more} and Sec. \ref{appendix:dataset}, respectively. 

\subsection{Results on Visual Perception}\label{sec:exp-perception}
Visual perception is an important ability of LVLMs. As presented in Sec. \ref{sec:method-capability}, we evaluate through image classification (ImgCls), multi-class identification (MCI), and object counting (OC). The evaluation details of tasks are provided in Appendix.~\ref{appendix:exp-perception}.
The evaluation results are reported in Table \ref{tab:oc_mci_results}. We have three observations. (1) mPLUG-Owl and LLaVA perform best on coarse-fined classification tasks (\textit{i.e.,} ImageNet1K and CIFAR10). The commonality is that they update LLM with 158K instruction-following data. (2) InstructBLIP presents good perception ability in fine-grained ImgCls, OC, and MCI tasks. The main reason is that InstructBLIP is fine-tuned on 1.6M VQA data, making it overfit on these tasks. (3) The performances of LVLMs on ImgCls are significantly inferior to supervised SOTA, indicating plenty of room for LVLM's perception ability.

\begin{table*}[t]
\centering
\small
\caption{Comparison of Zero-shot Performance for Large-scale Vision and Language Models (LVLMs) on OCR, KIE, and Image Captioning Tasks. Evaluation metrics include word accuracy for OCR datasets, entity-level F1 score for KIE datasets, and CIDEr score for image captioning datasets.}
\label{tab:vka_results}
\resizebox{\textwidth}{!}{%
\begin{tabular}{c|c|ccccccccc}
\toprule
\multicolumn{2}{c|}{Datasets} & BLIP2 & InstructBLIP & LA-V2 & LLaVA & MiniGPT-4 & mPLUG-Owl & Otter & VPGTrans & S-SOTA \\
\cmidrule(lr){1-2}\cmidrule(lr){3-11}
\multirow{13}{*}{OCR}
& IIIT5K      & \underline{80.17} & \textbf{83.90} & 36.30 & 31.57 & 25.00 & 25.30 & 17.57 & 62.87 & 99.2\cite{clip4str} \\
& IC13        & \underline{81.13} & \textbf{82.08} & 20.87 & 16.39 & 16.69 & 14.98 & 09.67 & 71.11 & 98.4\cite{parseq} \\
& IC15        & \underline{66.68} & \textbf{73.57} & 29.40 & 26.58 & 22.05 & 20.99 & 18.49 & 55.90 & 91.4\cite{clip4str} \\
& Total-Text  & \underline{68.31} & \textbf{71.51} & 30.93 & 24.51 & 18.65 & 20.63 & 14.81 & 54.76 & 90.5\cite{centripetaltext} \\
& CUTE80      & \underline{85.07} & \textbf{86.11} & 35.76 & 36.46 & 33.33 & 31.94 & 18.75 & 70.49 & 99.3\cite{clip4str} \\
& SVT         & \underline{85.78} & \textbf{86.86} & 20.40 & 18.55 & 15.46 & 14.37 & 10.51 & 72.02 & 98.3\cite{clip4str} \\
& SVTP        & \underline{77.34} & \textbf{80.93} & 31.01 & 27.44 & 20.31 & 20.78 & 19.22 & 64.50 & 97.2\cite{clip4str} \\
& COCO-Text   & \underline{53.62} & \textbf{58.25} & 20.94 & 18.05 & 11.86 & 12.88 & 11.30 & 36.98 & 81.1\cite{clip4str} \\
& WordArt     & \underline{73.66} & \textbf{75.12} & 38.98 & 35.87 & 31.90 & 31.90 & 21.05 & 62.34 & 72.5\cite{wordart} \\
& CTW         & \underline{67.43} & \textbf{68.58} & 18.13 & 16.73 & 14.95 & 13.87 & 10.05 & 52.80 & 88.3\cite{centripetaltext} \\
& HOST        & \underline{57.28} & \textbf{61.22} & 16.60 & 15.94 & 13.45 & 11.88 & 10.14 & 50.58 & 77.5\cite{clip4str} \\
& WOST        & \underline{68.83} & \textbf{73.26} & 21.73 & 20.49 & 19.12 & 14.65 & 12.29 & 57.66 & 87.5\cite{clip4str} \\
\cmidrule(lr){1-2}\cmidrule(lr){3-11}
\multirow{2}{*}{KIE}
& SROIE & \underline{0.08} & \textbf{0.09} & 0.02 & 0.01 & 0.02 & 0.01 & 0.01 & 0.02 & 97.81\cite{layoutlmv2} \\
& FUNSD & 1.02 & 1.03 & \textbf{2.16} & \underline{1.93} & 1.27 & 0.41 & 1.91 & 1.20 & 89.45\cite{geolayoutlm} \\
\cmidrule(lr){1-2}\cmidrule(lr){3-11}
\multirow{3}{*}{Image Captioning}
& NoCaps     & \textbf{48.58} & \underline{46.33} & 41.66 & 33.09 & 42.43 & 28.30 & 29.23 & 48.13 & 124.77\cite{git} \\
& Flickr-30k & \underline{46.48} & \textbf{50.45} & 30.49 & 27.65 & 26.04 & 20.53 & 23.00 & 32.51 & - \\
& WHOOPS & \underline{96.12} & \textbf{97.98} &  57.60 &  34.36 &  47.36 &  42.73 & 32.70 & 38.38 & - \\
\cmidrule(lr){1-2}\cmidrule(lr){3-11}
\multicolumn{2}{c|}{Average Score} & \underline{0.927} & \textbf{0.967} & 0.443 & 0.377 & 0.346 & 0.286 & 0.237 & 0.720 & - \\
\bottomrule
\end{tabular}%
1}
\end{table*}

\subsection{Results on Visual Knowledge Acquisition}
Visual knowledge acquisition involves going beyond image perception to acquire deeper understanding and knowledge. In our study, we evaluate the acquisition of visual knowledge through various tasks, namely Optical Character Recognition (OCR), Key Information Extraction (KIE), and Image Captioning, all performed in a Visual Question Answering (VQA) fashion. The evaluation details of tasks are demonstrated in Appendix.\ref{appendix:exp-vka}.
Table~\ref{tab:vka_results} shows the zero-shot performance in visual knowledge acquisition, and we have the following observations. First, BLIP2, InstructBLIP, and VPGTrans achieve dominant performance in all tasks. This may be because these models use a large visual encoder (i.e., ViT-g/14) and Q-Former updated with massive image-text pairs. A stronger visual encoder and adaption module can extract better tokens entailed with the global and local context, leading to remarkable improvement in visual knowledge acquisition. Second, InstructBLIP presents  consistently the best results on almost all tasks. The main reason is that InstructBLIP overfits these tasks by fine-tuning with massive VQA data.

\begin{table*}[t]
\centering
\small
\caption{Comparison of Zero-shot Performance for LVLM Models on VQA, KGID, and VE Tasks. For VQA and KGID tasks, Mean Reciprocal Rank (MRR) is used for the Visdial, while top-1 accuracy is employed for the remaining tasks.}
\label{tab:vr_results}
\resizebox{\textwidth}{!}{%
\begin{tabular}{c|c|ccccccccc}
\toprule
\multicolumn{2}{c|}{Datasets} & BLIP2 & InstructBLIP & LLaMA-Adapter-v2 & LLaVA & MiniGPT-4 & mPLUG-Owl & Otter & VPGTrans & S-SOTA \\
\cmidrule(lr){1-2}\cmidrule(lr){3-11}
\multirow{10}{*}{VQA}
& DocVQA  &  4.75 &  5.89 & \textbf{ 8.13} & \underline{ 6.26} &  2.65 &  2.24 &  3.44 &  3.53 & 54.48\cite{docvqa} \\
& TextVQA & 31.98 & \underline{39.60} & \textbf{43.76} & 38.92 & 19.40 & 38.76 & 21.52 & 21.98 & 73.1\cite{pali} \\
& STVQA   & 20.98 & 28.30 & \textbf{32.33} & \underline{28.40} & 13.55 & 12.10 & 15.23 & 17.13 & - \\
& OCR-VQA & \underline{38.85} & \textbf{60.20} & 38.12 & 23.40 & 16.85 &  8.84 & 19.50 & 21.71 & - \\
& OKVQA   & 44.93 & \textbf{60.52} & \underline{55.93} & 54.36 & 37.48 & 22.89 & 49.01 & 44.51 & - \\
& GQA     & \underline{45.53} & \textbf{49.96} & 43.93 & 41.30 & 30.82 & 14.02 & 38.12 & 32.99 & 72.1\cite{coarse} \\
& Visdial & 10.73 & \textbf{45.20} & 12.92 & \underline{14.66} & 10.31 & 13.34 & 11.67 & 9.70 & 68.92\cite{factor} \\
& IconQA  & \textbf{62.82} & \underline{56.25} & 41.83 & 42.95 & 37.59 & 11.64 & 26.77 & 38.22 & 83.62\cite{iconqa} \\
& VSR     & \textbf{63.63} & 41.28 & 50.63 & \underline{51.24} & 41.56 & 24.74 & 06.40 & 48.77 & 70.1\cite{vsr} \\
& WHOOPS      & \underline{24.87} & \textbf{30.13} & 24.15 & 24.39 & 17.91 & 20.70 & 15.14 & 15.88 & - \\
\cmidrule(lr){1-2}\cmidrule(lr){3-11}
\multirow{2}{*}{KGID}
& ScienceQA IMG & \textbf{60.73} & 46.26 & \underline{54.19} & 49.33 & 25.43 &  2.80 & 27.22 & 36.99 & 92.53\cite{llava} \\
& VizWiz        & \textbf{65.44} & \underline{65.31} & 62.07 & 62.42 & 47.48 & 38.99 & 50.04 & 53.23 & 73.3\cite{pali} \\
\bottomrule
\cmidrule(lr){1-2}\cmidrule(lr){3-11}
VE & SNLI-VE & 32.00 & \textbf{59.00} & \underline{58.8} & 57.80 & 54.80 & 54.50 & 52.60 & 52.20 & - \\
\cmidrule(lr){1-2}\cmidrule(lr){3-11}
\multicolumn{2}{c|}{Average Score} & 0.759 & \textbf{0.908} & \underline{0.833} & 0.771 & 0.527 & 0.420 & 0.513 & 0.588 & - \\
\bottomrule
\end{tabular}%
}
\end{table*}

\subsection{Results on Visual Reasoning}
Visual reasoning encompasses the ability to comprehensively understand images and perform cognitive tasks. In this section, we evaluate the visual reasoning ability of LVLMs on various tasks, including Visual Question Answering (VQA), Knowledge-Grounded Image Description (KGID), and Visual Entailment (VE) tasks. The evaluation details of tasks are provided in Appendix.\ref{appendix:exp-reasoning}.
Table~\ref{tab:vr_results} shows the zero-shot performance in visual reasoning, and we have the following observations. First, compared with BLIP2, InstructBLIP again presents  better results overall because it overfits many tasks by fine-tuning massive VQA data. Second, compared with BLIP2, instruction-tuned LVLMs, except for InstructBLIP, generally perform worse than BLIP2. The common words in the instruction data often influence the generated content, which can not be evaluated by the current metrics (see Fig. \ref{fig:CIDEr-failure-case}). Third, instruction-tuned LVLMs consistently surpass BLIP2 on SNLI-VE where the final answer is obtained by multi-turn reasoning. It shows that instruction-following fine-tuning can produce promising content once a good evaluation scheme is employed. We provide more evidence in Fig. \ref{appfig:multi-turn-pope} in Sec. \ref{appendix:more} of Appendix.




\begin{table*}[t]
\centering
\small
\label{tab:vc_results}
\resizebox{\textwidth}{!}{%
\begin{tabular}{c|c|ccccccccc}
\toprule
\multicolumn{2}{c|}{Datasets} & BLIP2 & InstructBLIP & LA-v2 & LLaVA & MiniGPT-4 & mPLUG-Owl & Otter & VPGTrans & S-SOTA \\
\cmidrule(lr){1-2}\cmidrule(lr){3-11}
\multirow{6}{*}{ImageNetVC}
& Color  & 26.22 & \textbf{67.78} & 36.12 & \underline{43.70} & 24.49 & 26.20 & 26.21 & 23.34 & 44.70\cite{xia2023imagenetvc} \\
& Shape  & 34.21 & \textbf{59.06} & 28.63 & \underline{39.10} & 23.54 & 34.19 & 34.20 & 23.92 & 40.50\cite{xia2023imagenetvc} \\
& Mater. & 35.79 & \underline{63.50} & 33.86 & \textbf{65.58} & 28.56 & 35.82 & 35.81 & 27.26 & 61.90\cite{xia2023imagenetvc} \\
& Compo. & 50.71 & \textbf{83.25} & 50.13 & 56.73 & \underline{59.26} & 50.73 & 50.72 & 56.43 & 54.00\cite{xia2023imagenetvc} \\
& Others & 34.48 & \textbf{68.37} & 32.69 & \underline{59.38} & 39.38 & 34.50 & 34.49 & 35.83 & 51.70\cite{xia2023imagenetvc} \\
\cmidrule(lr){1-2}\cmidrule(lr){3-11}
\cmidrule(lr){1-2}\cmidrule(lr){3-11}
VCR & VCR & 31.60 & \textbf{54.20} & \underline{49.80} & 48.20 & 49.00 & 46.00 & 47.00 & 41.40 & - \\
\cmidrule(lr){1-2}\cmidrule(lr){3-11}
\multicolumn{2}{c|}{ Average Score} & 0.535 & \textbf{0.995} & 0.589 & \underline{0.791} & 0.565 & 0.579 & 0.582 &  0.522 & - \\
\bottomrule
\end{tabular}%
}
\caption{Comparisons of Zero-shot visual commonsense Performance for LVLM Models on VCR and ImageNetVC datasets. Top-1 accuracy is employed for the two datasets.}
\label{tab: visual commonsense}
\end{table*}

\subsection{Results on Visual Commonsense}
The visual commonsense evaluation aims to evaluate the model's comprehension of commonly shared human knowledge about generic visual concepts. We use two challenging visual commonsense benchmarks in a zero-shot setting, including ImageNetVC and Visual Commonsense Reasoning (VCR). The evaluation details of tasks are presented in Appendix.\ref{appendix:exp-commonsense}.

As shown in Table~\ref{tab: visual commonsense}, we can find that all those LVLMs can partly solve visual commonsense problems. First, InstructBLIP performs best among those LVLMs on the ImageNetVC and VCR dataset. The main reason is that it is fine-tuned on 1.6M fine-grained VQA data, making it adapt to answer visual common questions. Second, LLaVA also performs well in the visual commonsense task. The reason is that it updates LLM with instruction-following data. Third, instruction-tuned LVLMs again surpass BLIP2 on the two Visual Commonsense tasks.  It shows that instruction-tuning can provide more effective clues than BLIP2 for visual commonsense. Note that the final answer of VCR is obtained by multi-turn reasoning. It also shows the significant role of a good evaluation scheme in producing promising content for instruction-tuned models.


\subsection{Results on Object Hallucination}
Although LVLMs have made significant progress, they still struggle with the issue of hallucination, which refers to their tendency to produce objects that do not align with the descriptions provided in the target images. In this section, we focus on evaluating such object hallucination problems on MSCOCO captioning dataset. Following POPE \cite{li2023evaluating} evaluation pipeline which is a multi-step QA procedure, we prompt LVLMs with multiple Yes-or-No questions. Each image is prompted with $6$ Yes-or-NO questions. For example, `\textit{Is there a person in the image?}'. We use accuracy, precision, recall, F1-Score and the ratio of answering `Yes' as evaluation metrics. The evaluations are three datasets including MSCOCO-Random/Popular/Adversarial \cite{li2023evaluating}. From Random to Adversarial, the questions become more and more difficult.  
%
From Table~\ref{tab:pope_results}, we could come to the following conclusions. InstructBlip performs best in the hallucination problem, followed by BLIP2, whose average accuracy reacheS more than 80\%.
We find that instruction-tuned models, except for InstructBLIP, perform worse than BLIP2 because they tend to answer `Yes' to the question, which shows that LVLMs are prone to generate objects which do not exist in the image. In sec. \ref{appsec:exp-object-hallucination}, we show that such an object hallucination problem can be alleviated by a multi-turn reasoning pipeline, which can be also seen from  the experiments on SNLI-VE and VCR.

\begin{table*}[t]
\centering
\small
\caption{Detailed evaluation results of the zero-shot performance of LVLMs on MSCOCO using POPE evaluation pipeline \cite{li2023evaluating}, where accuracy represents the accuracy of prediction; precision represents how many of the predicted positive samples are true positive samples; recall represents how many of all true positive samples are correctly identified; and yes represents the probability that the model outputs a yes answer. The average score is calculated based on the metric of accuracy.}
\label{tab:pope_results}
\resizebox{\textwidth}{!}{%
\begin{tabular}{c|c|cccccccc}
\toprule
Datasets & Metrics & BLIP2 & InstructBLIP & LA-V2 & LLaVA & MiniGPT-4 & mPLUG-Owl & Otter & VPGTrans \\
\cmidrule(lr){1-2}\cmidrule(lr){3-10}
\multirow{5}{*}{MSCOCO-Random}
& Accuracy  & \underline{82.21} & \textbf{88.83} & 74.44 & 51.52 & 52.58 & 61.37 & 61.40 & 48.28 \\
& Precision & 97.48 & 96.01 & 68.24 & 51.54 & 68.63 & 57.89 & 57.82 & 74.17 \\
& Recall    & 67.27 & 81.60 & 94.00 & 100.00 & 57.50 & 97.52 & 95.92 & 56.78 \\
& F1-Score  & \underline{79.61} & \textbf{88.23} & 79.08 & 68.03 & 62.57 & 72.65 & 72.15 & 64.32 \\
& Yes       & 35.58 & 43.99 & 70.99 & 100.00 & 44.25 & 87.15 & 85.76 & 47.38 \\
\cmidrule(lr){1-2}\cmidrule(lr){3-10}
\multirow{5}{*}{MSCOCO-Popular}
& Accuracy  & \underline{80.10} & \textbf{84.15} & 56.82 & 50.00 & 49.31 & 55.83 & 49.56 & 47.86 \\
& Precision & 90.49 & 85.96 & 53.89 & 50.00 & 63.56 & 53.61 & 50.07 & 70.37 \\
& Recall    & 67.27 & 81.60 & 94.20 & 100.00 & 58.03 & 97.13 & 95.92 & 55.90 \\
& F1-Score  & \underline{77.17} & \textbf{83.72} & 68.56 & 66.67 & 60.67 & 69.09 & 65.79 & 62.92 \\
& Yes       & 37.17 & 47.47 & 87.40 & 100.00 & 48.29 & 91.20 & 96.58 & 51.92 \\
\cmidrule(lr){1-2}\cmidrule(lr){3-10}
\multirow{5}{*}{MSCOCO-Adversarial}
& Accuracy  & \underline{78.52} & \textbf{81.95} & 60.52 & 50.00 & 49.62 & 54.43 & 50.68 & 47.86 \\
& Precision & 86.83 & 82.05 & 54.58 & 50.00 & 62.55 & 52.73 & 50.56 & 69.76 \\
& Recall    & 67.27 & 81.60 & 96.45 & 100.00 & 58.71 & 97.59 & 95.92 & 59.22 \\
& F1-Score  & \underline{75.81} & \textbf{81.82} & 69.12 & 66.67 & 68.47 & 72.09 & 66.22 & 64.06 \\
& Yes       & 38.73 & 49.77 & 88.23 & 100.00 & 48.54 & 92.95 & 95.31 & 52.27 \\
\cmidrule(lr){1-2}\cmidrule(lr){3-10}
\multicolumn{2}{c|}{ Average Score} & \underline{0.945} & \textbf{1.00} & 0.751 & 0.595 & 0.594 & 0.673 & 0.633 & 0.565 \\
\bottomrule
\end{tabular}%
}
\end{table*}


\begin{figure*}[t]
		\centering
		\includegraphics[width=1.0\linewidth]{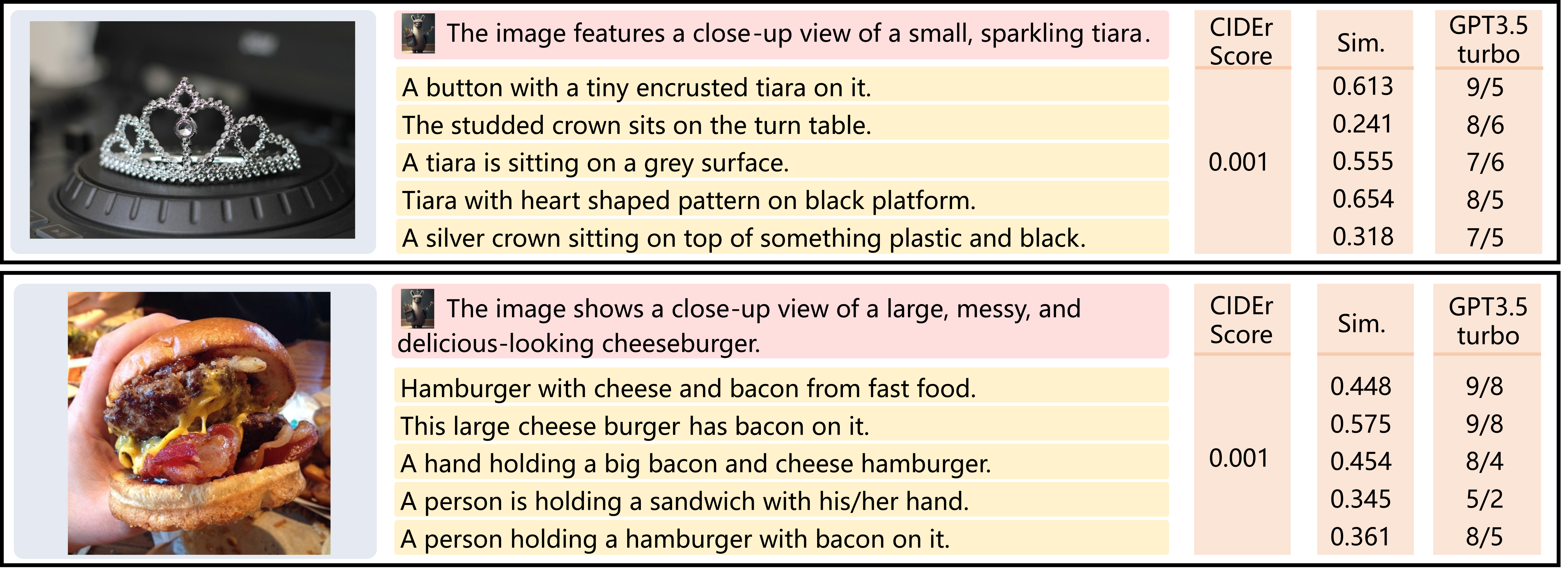}
		\caption{\textbf{Limitations of CIDEr Score Evaluation in Image Captioning}. SentenceTransformer~\cite{SentenceTransformer} computes the similarity between generated and ground-truth text as "Sim." ``GPT3.5 Turbo'' indicates that we feed GPT with the most elaborate ground-truth caption and use GPT to evaluate the quality of the generated captions (left score) and other ground-truth captions (right score). The prompt we used is similar to Vicuna's GPT-4 evaluation template~\cite{vicuna}.}
		\label{fig:CIDEr-failure-case}
\end{figure*}

\subsection{Results on Embodied Intelligence}\label{sec:exp-embodied}
In this section, we present the evaluation results focusing on embodied intelligence. To appraise the effectiveness of planning outputs using the given image, we conducted a user study involving 15 participants. The study comprised 6 household scenarios carefully selected from VirtualHome~\cite{puig2018virtualhome}. Specifically, the participants rated the generated plans from different LVLM models using a scoring system similar to~\cite{embodiedgpt}. The evaluation comprised five dimensions with scores ranging from 1 to 5. These dimensions included object recognition accuracy, spatial relationship understanding, level of conciseness in the response, reasonability of the planning, and executability of the planning. The resulting average scores for the different models among the participants are presented in Table~\ref{tab:embodied_tasks_eval} below. Furthermore, in the Appendix C, we present quantitative evaluation results for Franka Kitchen~\cite{gupta2019relay}, Minecraft~\cite{fan2022minedojo}, and Meta-World~\cite{yu2020meta}.
Based on the findings, two deductions can be made. Firstly, the use of image-text pairs is consequential in aligning visual-text features. This is evident from the comparison demonstrated in Table~\ref{tab:comparison-LVLMs}. Otter and InstructBLIP lacked the training process on image-text pairs which necessitates the alignment of visual reasoning and text description, resulting in a degraded ability of spatial relationship analysis. Conversely, mPLUG-Owl outperformed Otter partially due to 204M image-text training pairs. Secondly, we observe that visual instruction data is essential for embodied tasks. BLIP2 lacked visual instruction tuning, which greatly affected its capability of producing reasonable and executable plans.

%

\begin{table*}[t]
\centering
\small
\caption{Generated planning quality evaluation on embodied tasks.} 
\label{tab:embodied_tasks_eval}
\resizebox{\textwidth}{!}{%
\begin{tabular}{c|c|ccccccccc}
\toprule
\multicolumn{2}{c|}{Dataset} & BLIP2 & InstructBLIP & LA-V2 & LLaVA & MiniGPT-4 & mPLUG-Owl & Otter & VPGTrans  \\
\cmidrule(lr){1-2}\cmidrule(lr){3-10}
\multirow{5}{*}{VirtualHome}
& Object Recon.($\uparrow$)  & 2.03 & 3.08 & \underline{3.81} & \textbf{3.88} & 3.70 & 3.42 & 3.38 & 3.43   \\
& Spatial Relation.($\uparrow$)  &  1.68 &2.78 &	\textbf{3.71} &	\underline{3.61} & 3.47 &	3.22 &	3.10 &	3.22 \\
& Conciseness ($\uparrow$) & \textbf{3.25} &	\underline{2.48} &	2.04 &	1.86 &	1.62 &	1.48 &	1.86 	& 1.76  \\
& Reasonability($\uparrow$) & 2.78 	& 3.20 &	\textbf{4.04} &	\underline{3.70} &	3.54 &	3.44 &	3.07 	& 3.35  \\
& Executability($\uparrow$) & 2.88 	& 3.10 &	\textbf{4.08} &	\underline{3.82} &	3.11 &	3.54 &	3.12 	& 3.35  \\
\cmidrule(lr){1-2}\cmidrule(lr){3-10}
\multicolumn{2}{c|}{Average Score} & 0.674 & 0.772 & \textbf{0.922} & \underline{0.879} & 0.805 & 0.785 & 0.761 & 0.789 \\
\bottomrule
\end{tabular}%
}
\end{table*}

\subsection{Results on Online Arena Evaluation}\label{sec:exp-arena}
The arena features anonymous and randomized pairwise battles in a crowd-sourced manner. We have collected $634$ and $1425$ pieces of evaluation data up until June 3 and June 13 in 2023, respectively. The collected data shows almost the same number of battle outcomes for `Model A wins' and `Model B wins.' Moreover, $21.8$\% and $22.0$\% battle outcomes are voted as `both are bad' in two copies of collected data, respectively, implying that the current LVLMs still struggle to generate good answers for open-world visual questions. Furthermore, we rank the selected $8$ LVLMs with Elo rating \cite{elo1967proposed} using two copies of the collected data by following Fastchat \cite{zheng2023fastchat} and \cite{bai2022training}. As shown in Fig. \ref{fig:lvlm-radar} (b) and (c), mPLUG-Owl, MiniGPT-4, LLaMA-Adapter V2, Otter, and VPGTrans, which are fine-tuned with amounts of instruction-following data with updating many parameters, are the top-3 best models in the open-world VQA scenario according to two ranking lists, indicating the significance of instruction-following tuning and effective parameter update. Moreover, InstructBLIP perform best on in-domain capability evaluation, while being much worse than many instruction-tuned models, implying severe overfitting issue, as shown in Fig. \ref{fig:lvlm-radar}.

\section{Discussion and Conlcusion}

\textbf{New Evaluation Metrics.}
Our quantitative evaluation mainly uses the CIDEr score and accuracy. The CIDEr score is widely used in image captioning and QA evaluation. It measures the similarities between generated and ground-truth answers. However, LVLMs' responses are diverse, in different styles with the ground truth. As such, the CIDEr score is unsuitable (see Fig. \ref{fig:CIDEr-failure-case} for failure cases). 
We also tried model-based evaluation, which uses Sentence Transformer \cite{SentenceTransformer} to calculate the feature similarity between generated and the ground-truth answers. It is generally more robust but sometimes suffers due to model limitations. 
Recent studies use the powerful Chat-GPT or GPT-4 as a judge to evaluate LLMs' responses. However, in LVLM evaluation, GPT is blind to the image and is inaccurate in some cases. 
We propose LVLM Arena, an innovative evaluation framework that utilizes a 1v1 LVLM battle with human judgment for open-world evaluation, leading to more accurate and realistic evaluation results. However, it requires significant human effort to produce reliable rating results, especially when numerous models exist. 
Therefore, developing fast, accurate, and generalized evaluation metrics for LVLMs remains an open problem.

\textbf{A Platform for LVLM Evaluation.}
We have developed an evaluation framework aimed at comprehensively assessing the performance of LVLM models across six critical capabilities. Each capability encompasses multiple tasks, with several datasets incorporated into each task. Our user-friendly interface allows users to contribute their own datasets and models, facilitating a collaborative and inclusive environment. With just one click, users can effortlessly access a holistic assessment of their target LVLM model through our evaluation platform. We are dedicated to regularly updating the datasets and expanding our support for a wider range of LVLM models on our platform. Users are encouraged to contribute their LVLM models by utilizing our platform's model inference interface. 
Additionally, we offer free online inference services for the LVLM models supported by LVLM Arena. This arena not only allows users to vote for their preferred models but also provides an Elo rating rank system that incorporates valuable human feedback, ensuring continuous improvement and refinement.

\textbf{Conclusion.}
This paper proposes a comprehensive evaluation benchmark for large vision-language models called LVLM-eHub that incorporates both quantitative performance evaluation and human feedback evaluation. For the quantitative evaluation, we employ 16 tasks spanning over 40+ text-related visual datasets to assess the six essential capabilities of LVLM models. Additionally, we have established an online LVLM Arena to gather human feedback on LVLM models continually. This arena serves as an invaluable resource, providing an Elo rating rank that offers LVLMs ranking in the open-world scenario.
Our evaluation results reveal several important findings, stimulating the future development of LVLMs.

\newpage

\bibliographystyle{unsrt}
\bibliography{egbib}



\newpage
\appendix

\renewcommand{\thefigure}{A.\arabic{figure}}
\setcounter{figure}{0}

\renewcommand{\thetable}{A.\arabic{table}}
\setcounter{table}{0}

\begin{figure*}[t]
		\centering
		\includegraphics[width=0.8\linewidth]{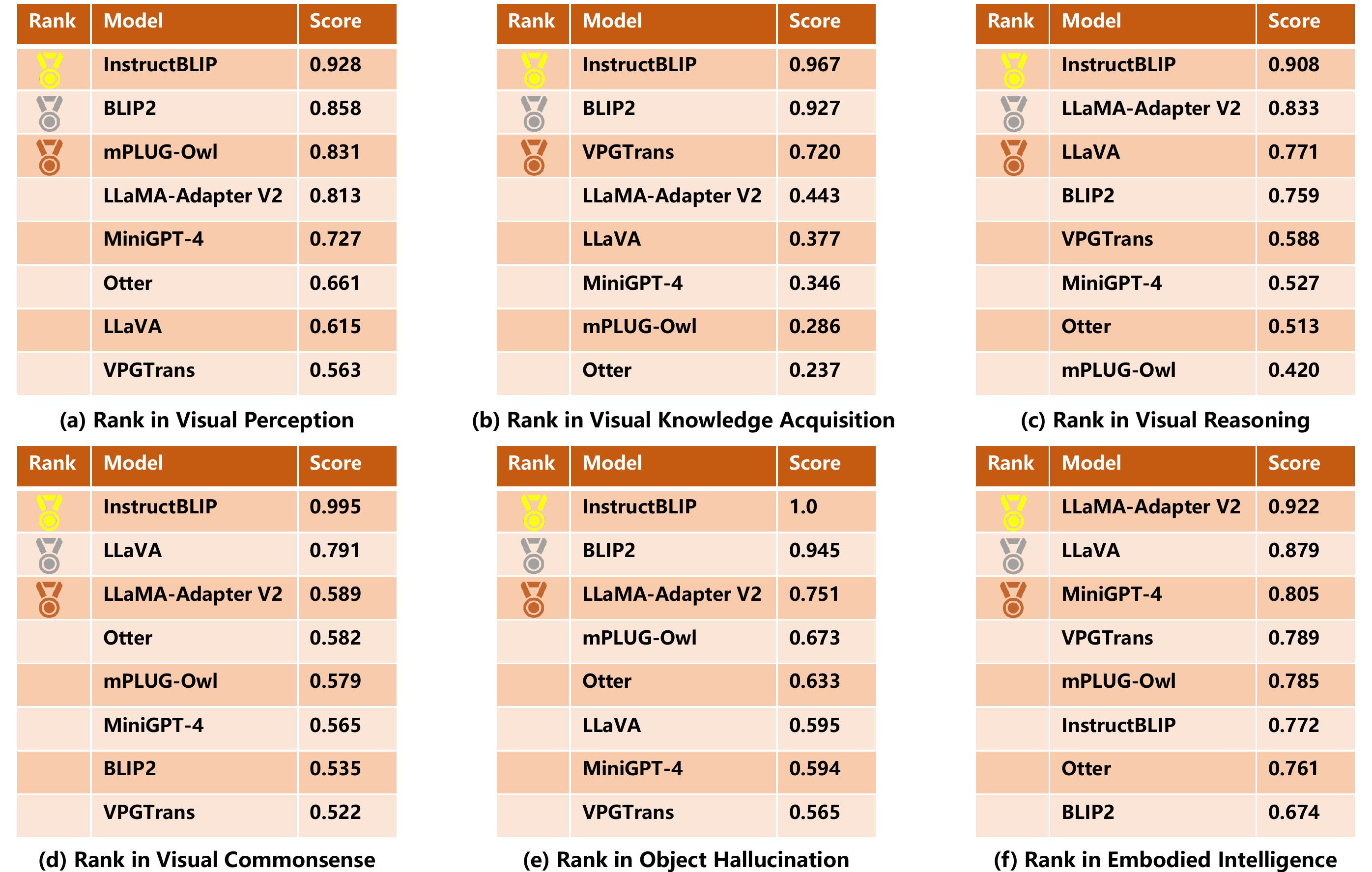}
		\caption{Comparison of LVLMs in the LVLM-eHub. (a - f) present the ranking list of LVLMs in terms of $6$ categories of multimodal capabilities, respectively, including visual perception, visual knowledge acquisition, visual reasoning, visual commonsense, object hallucination, and embodied intelligence.  }
		\label{fig:lvlm-radar-list}
\end{figure*}

In the appendix, we provide more details of LVLM-eHub and task settings of evaluation in Sec.~\ref{appendix:lvm_model_hub} and Sec.~\ref{appendix:exp}, respectively. Additionally, more experiments are illustrated in Sec.~\ref{appendix:more}. The evaluation datasets are summarized in Sec. \ref{appendix:dataset}.

\section{More details about our LVLM-eHub}\label{appendix:lvm_model_hub}
\subsection{Overall Evaluation Results}

%

\textbf{Our Findings.} We present our observations from extensive evaluation experiments in the following.
\begin{itemize}
    \item \textit{Instruction-tuned LVLM with massive in-domain data such as InstructBLIP heavily
overfits many existing tasks, generalizing poorly in the open-world scenario.} As shown in Fig. \ref{fig:lvlm-radar} and Fig. \ref{fig:lvlm-radar-list}, InstructBLIP achieves the best results in 5 categories of capabilities while lagging behind other instruction-tuned models such as LLaMA-Adapter V2 and mPLUG-Owl in embodied AI and LVLM arena platform. We see that InstructBLIP is fine-tuned on 16M visual question answering pairs (see Table \ref{tab:comparison-LVLMs}, exhibiting in many in-domain tasks such as perception and reasoning tasks. However, Embodied AI tasks require that the model is capable of generating a step-by-step plan for instruction with a given image. Moreover, the arena platform evaluates LVLMs' ability of visual question answering in open-world scenarios. InstructBLIP overfits in-domain tasks, generalizing poorly in these two real-world tasks.
\item \textit{ Instruction-tuned LVLM with moderate high-quality instruction-following data may result
in object hallucination issues.} The issue means that LVLMs would generate objects that are inconsistent with target images in the descriptions. It either makes the current evaluation metric such
as CIDER for image captioning ineffective or generates wrong answers. For instance, LLaMA-Adapter V2 can generate high-quality image captions which yet present a low CIDEr score as shown in Fig. \ref{fig:CIDEr-failure-case}. But the high sentence similarities between the generated answer and ground-truth answers measured by Sentence Transformer \cite{SentenceTransformer} and GPT3.5 shows that the generated answer is relatively accurate. Therefore, the instruction-tuned models could generate content that cannot be evaluated by existing metrics. It also indicates that it is urgent to develop an effective metric for LVLM evaluation. 

\,\,\,\, In addition, we also find that instruction-tuned LVLMs with moderate high-quality data are more likely to generate wrong answers. As shown in Table \ref{tab:pope_results}, LLaMA-Adapter V2, LLaVA, MiniGPT-4, mPLUG-Owl, Otter, and VPGTrans generally present higher accuracy and recall, and lower precision than BLIP2 and InstructBLIP. These models are tuned with moderate high-quality data such as LLaVA-158K or instruction-following data generated by LLM as shown in Table \ref{tab:comparison-LVLMs}. This implies that instruction-tuned LVLMs with moderate high-quality data are prone to answer `Yes' regardless of the accuracy of the answer to the underlying question.

\item \textit{Employing a multi-turn reasoning evaluation framework can mitigate the issue of
object hallucination, shedding light on developing an effective metric for LVLM evaluation.} In Table \ref{tab:vr_results} and Table \ref{tab: visual commonsense}, we see that instruction-tuned LVLMs with moderate high-quality data can achieve better performance than BLIP on SNLI-VE and VCR tasks under a multi-turn reasoning evaluation pipeline in Sec. \ref{sec:evaluation methods}. We also provide more evidence to demonstrate the effectiveness of such an evaluation technique in mitigating object hallucination in Fig. \ref{appfig:multi-turn-pope}.

\end{itemize}

\subsection{Model Details in LVLM-eHub}
\begin{itemize}
    \item \textbf{BLIP2} \cite{li2023blip} pre-trains a lightweight Q-Former on 129M image-text pairs. It follows a two-stage strategy to bridge the modality gap. The first stage bootstraps vision-language representation learning from a frozen image encoder ViT-g/14 in EVA-CLIP \cite{fang2023eva}. The second stage bootstraps vision-to-language generative learning from a frozen LLM FlanT5-XL \cite{chung2022scaling}, which enables zero-shot instructed image-to-text generation.
    \item \textbf{LLaVA} \cite{liu2023visual} connects the visual encoder ViT-L/14 of CLIP \cite{radford2021learning} with the language decoder LLaMA \cite{touvron2023llama} by a lightweight fully-connected (FC) layer. LLaVA first trains the FC layer with 595K image-text pairs while freezing the visual encoder and LLM and then fine-tunes the FC layer and LLM on 158K instructional vision-language data. 
    \item \textbf{LLaMA-Adapter V2 (LA-V2)} \cite{gao2023llama} is a parameter-efficient visual instruction model. Although the visual encoder (ViT-L/14) and LLM are kept frozen, LLaMA-Adapter V2 distributes the instruction-following ability of the whole LLaMA through bias(B)-tuning. In this way, the scale, bias, norm, and prompt parameters are tuned on 200M image captioning data, 158K visual instruction-following data, and 52K language instruction-following data constructed by GPT-4 \cite{peng2023instruction}.
    \item \textbf{MiniGPT-4} \cite{zhu2023minigpt} connects the visual encoder and text encoder by an FC layer. It also first trains the FC layer with 5M image-text pairs and then fine-tunes it on 3.5K instructional vision-language data. Despite the simplicity, MiniGPT-4 needs to load a pretrained vision encoder of BLIP2 and Vicuna LLM \cite{vicuna}.
    \item \textbf{mPLUG-Owl} \cite{ye2023mplug} incorporates a visual abstractor, essentially the same as \textit{Perceiver Resampler} in Flamingo~\cite{alayrac2022flamingo}, to bridge pretrained visual encoder ViT-L/14 and LLM (LLaMA) with a two-stage finetuning procedure. It firstly fully finetunes both the visual encoder and visual abstractor on 204M image-text pairs. Then for the second stage, 158K LLaVA-Instruct data is utilized to parameter-efficiently finetune pretrained LLM via LoRA.
    \item \textbf{Otter} \cite{li2023otter} is a multimodal model with in-context instruction tuning based on OpenFlamingo \cite{alayrac2022flamingo} which comprises a LLaMA-7B language encoder and a CLIP ViT-L/14. Although the visual and text encoder are frozen, Otter trains extra 1.3B parameters  coming from adaption modules on 158K instruction-following data.
    \item \textbf{InstructBLIP} \cite{dai2023instructblip} is initialized from a pre-trained BLIP-2 model consisting of a ViT-g/14 image encoder, a Vicuna LLM and a Q-Former to bridge those two. During vision-language instruction tuning, only Q-Former is fine-tuned on 13 visual question-answering datasets.
    \item \textbf{VPGTrans} \cite{zhang2023vpgtrans} is a simple transferring technique that adapts a smaller LLM to a larger LLM.  It transfers the VPG of BLIP-2 (i.e. ViT-g/14) from  OPT6.7B to Vicuna7B by training Q-Former on 13.8M Image-Text pairs. In addition, the VPG and projector are further tuned on
MiniGPT-4’s 3.5K self-instruct data instances.
\end{itemize}


\section{Evaluation Details}\label{appendix:exp}

\subsection{Details of Visual Perception}\label{appendix:exp-perception}
\textbf{For ImgCls}, we test LVLMs on two coarse-grained benchmarks (\textit{i.e.,} ImageNet1K and CIFAR10) in top-1 accuracy and two fine-grained benchmarks (\textit{i.e.,} Pets37 and Flowers102) in per-class accuracy. Following KOSMOS-1 \cite{huang2023language}, the default prompt `\textit{The photo of the}' is used for all LVLMs for a fair comparison over coarse-grained benchmarks, while it is too general for fine-grained visual perception. When confronted with intricate fine-grained image classification tasks in a zero-shot manner, contemporary multi-modal models often encounter difficulties in accurately generating precise subclass names. To gain a deeper understanding of their capabilities and enable effective comparisons between them, we have heuristically designed prompts ``\textit{What is the specific category of the dog or cat in the image?}'' for the Pets37 dataset, and ``\textit{What is the specific category of the flower in the image?}'' for the Flowers102 dataset, respectively. Furthermore, the generated coherent sentence-style responses deviate from the standard image classification benchmark. To accommodate this discrepancy, we considered the prediction as correct if the model output contains the correct class name, which is inspired by MultiModal OCR\cite{liu2023hidden}.

\textbf{For OC} task, we test LVLMs on MSCOCO and VCR1.0~\cite{zellers2019recognition}. It involves querying the model about the number of objects belonging to an image's specific class of interest. To this end, we use the prompt `\textit{Question: How many [obj] are there in the image? Answer:}'. The generated answer is then compared with the ground truth. We report accuracy by treating OC as a classification problem.

\textbf{For MCI} task, we also test LVLMs on MSCOCO and VCR1.0~\cite{zellers2019recognition}. We ask the model to determine whether a certain object is present or absent by prompting `\textit{Question: Does [obj] exist in the image. Answer:}'. We also report the accuracy by treating MC as a Yes or No classification problem.


\subsection{Details of Visual Knowledge Acquisition}\label{appendix:exp-vka}
\textbf{For OCR} task, we test the selected LVLMs with twelve representative OCR datasets, which are inclusive of IIIT5K\cite{ittt5k}, ICDAR 2013(IC13)\cite{ic13}, ICDAR 2015 (IC15)\cite{ic15}, Total-Text\cite{total-text}, CUTE80\cite{cute80}, Street View Text (SVT)\cite{svt}, SVTP-Perspective (SVTP)\cite{svtp}, COCO-Text\cite{coco-text}, WordArt\cite{wordart}, SCUT-CTW1500 (CTW)\cite{ctw}, heavily occluded scene text (HOST)\cite{ost}, weakly occluded scene text (WOST)\cite{ost}. These benchmarks consist of a diverse range of images containing textual information which can make an adequate comparison between LVLMs. The performance of the LVLMs is compared with top-1 accuracy and the prompt we use is `\textit{what is written in the image?}'. 


\textbf{For KIE} task, we employ the SROIE\cite{sroie} and FUNSD\cite{funsd} benchmarks to evaluate LVLMs, which encompass diverse documents like receipts and forms that require specific information extraction. The performance of LVLMs is evaluated using entity-level F1 scores. Additionally, we utilize information-specific prompts for each piece of information that the model should extract. For instance, in the SROIE benchmark case, we use the prompt `\textit{what is the name of the company that issued this invoice?}' to extract company information and `\textit{where was this invoice issued?}' prompt for address information. 


\textbf{For ImgCap} task, 
we utilize two benchmarks, including NoCaps\cite{nocaps} and Flickr30K\cite{flickr30k}. Each benchmark provides a collection of images with corresponding captions. In evaluation, CIDEr scores are used to evaluate these models with the prompt `\textit{what is described in the image?}'.

\subsection{Details of Visual Reasoning}\label{appendix:exp-reasoning}
\textbf{For VQA} task, we utilize nine benchmarks: DocVQA\cite{docvqa}, TextVQA\cite{textvqa}, STVQA\cite{stvqa}, OCR-VQA\cite{ocr-vqa}, OKVQA\cite{okvqa}, GQA\cite{gqa}, IconQA\cite{iconqa}, Visual Spatial Reasoning (VSR)\cite{vsr}, and Visual Dialog (Visdial)\cite{visdial}. These benchmarks offer a diverse set of question-image pairs, covering a wide range of topics. The task requires LVLMs to not only understand the visual content but also comprehend and reason about the posed questions. For specific evaluation, we employ the Mean Reciprocal Rank (MRR) metric for Visdial and top-1 accuracy for the remaining datasets. These metrics provide insights into the model's ability to accurately answer questions across the various VQA benchmarks.

\textbf{For KGID} task, it evaluates the LVLM's capability to generate informative and accurate descriptions of images by incorporating external knowledge. To assess performance, we employ the ScienceQA\cite{scienceqa} and VizWiz\cite{vizwiz} benchmarks, which consist of images accompanied by textual descriptions and knowledge-based information. Notably, in the case of ScienceQA, we only utilize the samples that contain images.

\textbf{For VE} task, it evaluates the VLPM's capability to determine the logical relationship between image pairs. We employ the SNLI-VE [26] benchmark, which provides pairs of images along with corresponding textual premises and hypotheses. For efficient evaluation, we randomly select 500 samples from the dev split of the SNLI-VE dataset. We find that a naive QA pipeline is hard to give meaningful predictions. Therefore, as shown in Fig. \ref{fig:evalation-methods}, we employ multi-turn reasoning to solve SNLI-VE. There are three components in multi-turn reasoning pipeline: a Questioner, an Answerer, and a Reasoner. The Questioner first raises a set of sub-questions based on the main question, then Answerer produces the relative sub-answers, and Reasoner decide whether a confident answer is derived to its main question by analyzing the sub-questions and sub-answers. The "Questioner-Answerer-Reasoner" loop keeps iterating until the Reasoner derives a confident final answer or the number of iterations reaches a predefined maximum. Following You et al~\cite{you2023idealgpt}, three simple prompts are also applied to generate better sub-answers and sub-questions. We use ChatGPT as the Reasoner and Questioner via "gpt-3.5-turbo" API. The eight studied pre-trained LVLMs are served as the Answerers and produce image captions respectively for comparing their abilities to solve visual-entailment problem.

\subsection{Details of Visual Commonsense}\label{appendix:exp-commonsense}
\textbf{For ImageNetVC}, we evaluate the zero-shot visual commonsense of LVLMs. It contains high-quality QA pairs for various commonsense, including color, shape, mater, comp, and others. Specifically, as shown in Fig. \ref{fig:evalation-methods}, QA pairs in ImageNetVC~\cite{xia2023imagenetvc} are first transformed into prompts like `\textit{[Question] The answer is [Answer].}', and then each prompt is converted into a sequence of tokens. Secondly, the image and text tokens are transformed into a sequence of visual embeddings and a sequence of text embeddings, respectively. Finally, visual embeddings are prefixed into the text embeddings yielding the final embeddings which are put into a frozen pre-trained LVLM to calculate the score. The probability distribution over all answer candidates using softmax is calculated. We use the prefix-based score to choose the final answer with maximum likelihood. Following Xia et al\cite{xia2023imagenetvc}, five similar prompts are utilized to take average values for final evaluation among eight LVLMs. 

%

\textbf{For VCR}, it expects that the LVLMs can find the correct answer among four answer candidates. For efficient evaluation, we randomly select 500 samples from the val split of the VCR dataset. We find that a naive QA prompt cannot give meaningful output. Similar to the SNLI-VE evaluation (Section B.3), we adopt a multi-turn reasoning evaluation technique to solve the VCR task. 


\begin{figure*}[t]
		\centering
		\includegraphics[width=1.0\linewidth]{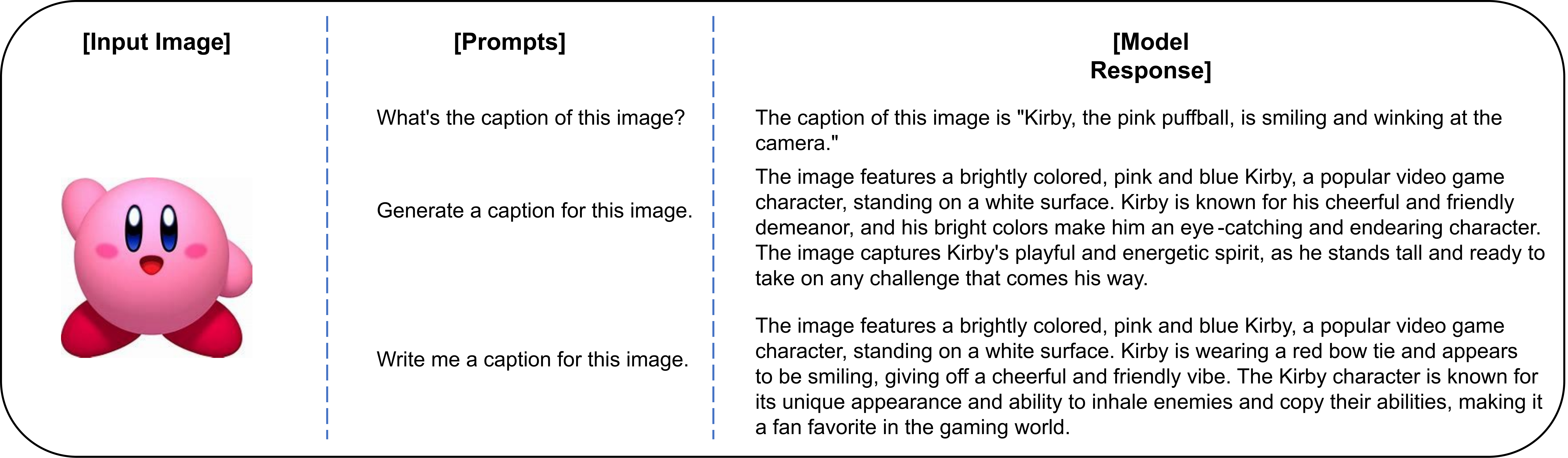}
		\caption{In this example, we present the variation in model responses of LA-V2 when provided with the same image but different prompts. All model responses mentioned here were generated using zero temperature and a top-p value of 0.75. The purpose is to showcase how the model's output changes based on the prompt input.}
		\label{fig:diff-prompts}
\end{figure*}

\begin{figure*}[htp]
\centering
\begin{subfigure}{0.6\textwidth}
\includegraphics[width=\textwidth]{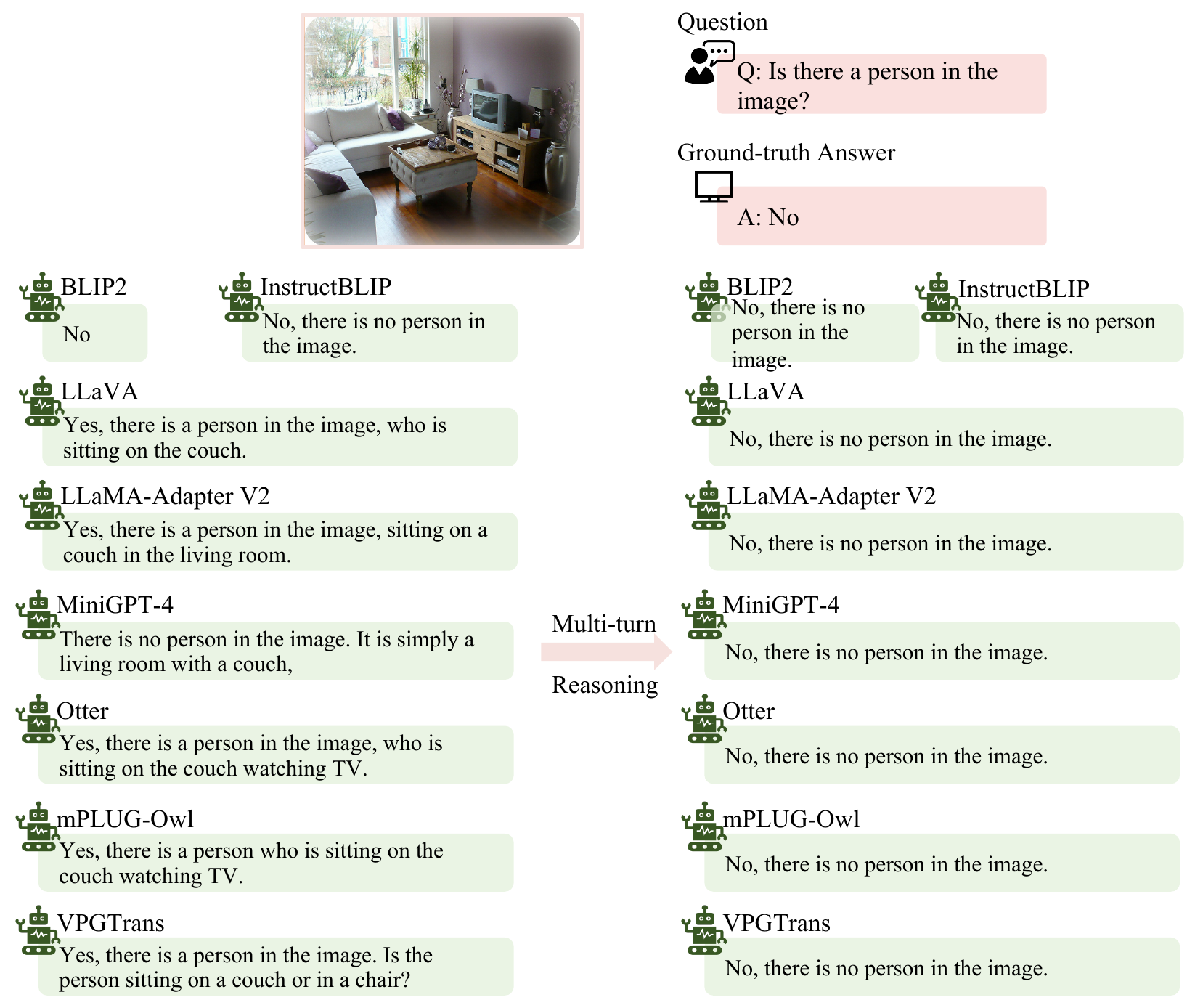}
\end{subfigure}

\bigskip
\begin{subfigure}{\textwidth}
\includegraphics[width=\textwidth]{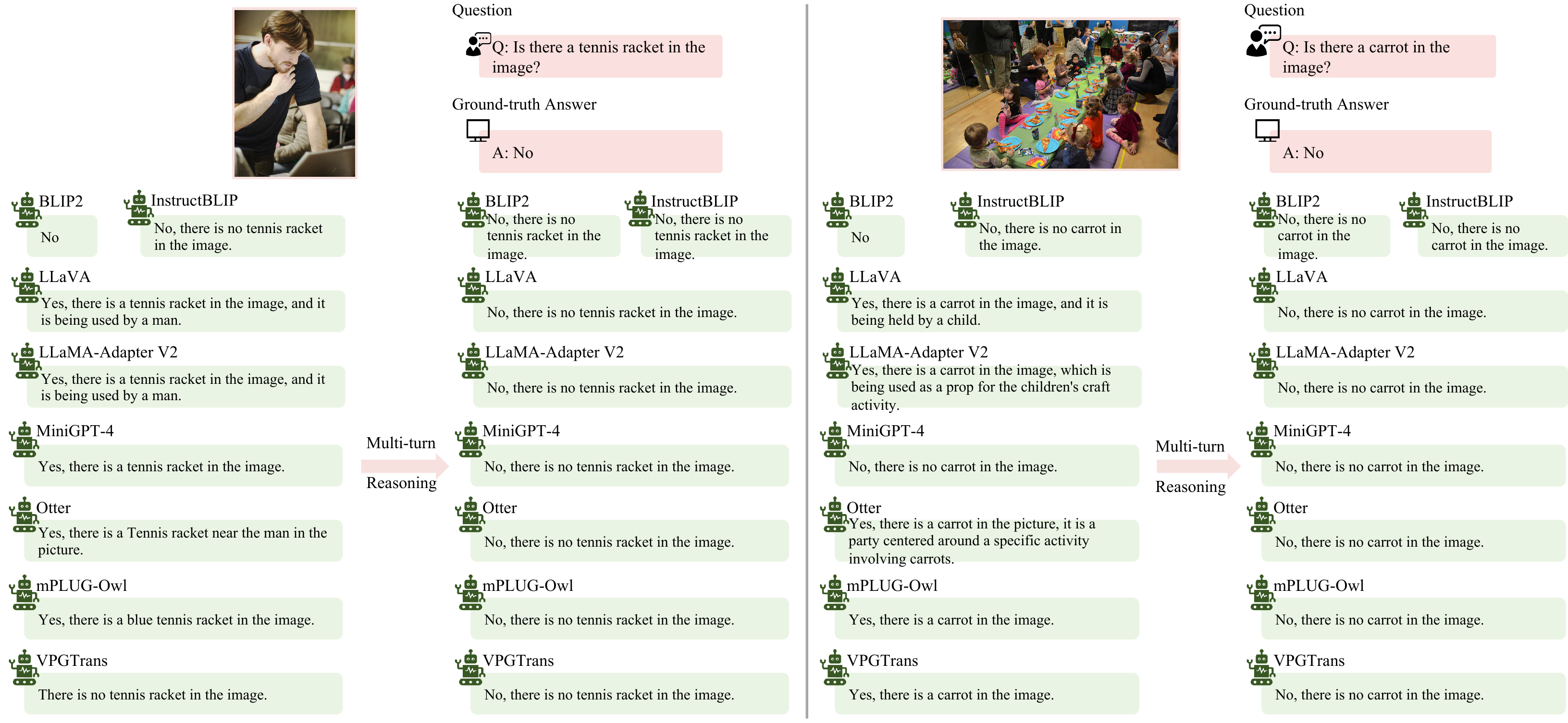}
\end{subfigure}

\caption{\textbf{The Effect of Multi-turn Reasoning Evaluation Pipeline}. We see that many LVLMs could generate content that does not exist in the given image, suffering from the object hallucination issue \cite{li2023evaluating}. Moreover, a multi-turn reasoning evaluation pipeline can mitigate object hallucination issues.}
\label{appfig:multi-turn-pope}
\end{figure*}

\section{More Experiments}\label{appendix:more}
\subsection{Sensitivity to Prompts.}
Throughout our comprehensive evaluation, we discovered that LVLM models are highly sensitive to the choice of prompts. An illustrative example of this sensitivity is observed in the image captioning task, where altering the prompt employed for the VPGTrans model on the NoCaps dataset leads to a substantial variation in performance, ranging from 19.66 to 48.13. Fig.~\ref{fig:diff-prompts} showcases some examples of the results generated based on different prompts. This sensitivity underscores the significance of carefully selecting and designing prompts to achieve optimal performance in LVLM-based tasks.

\subsection{More Results on Object Hallucination}\label{appsec:exp-object-hallucination}


\textbf{Multi-turn Reasoning Evaluation Can Alleviate Object Hallucination.} We also show the hallucination problem by visualizing some examples in Fig. \ref{appfig:multi-turn-pope}. It can be seen that LVLMs except for BLIP2 and InstructBLIP are prone to generate objects which are inconsistent with the image. Fortunately, such an issue can be mitigated by a multi-turn reasoning evaluation framework. We believe that the reasoning procedure can encourage instruction-tuned models to re-organize the knowledge they grasp and finally generate the right answers. It is significant to investigate how to evaluate instruction-tuned LVLMs in the right way.

\subsection{More Results on Embodied Tasks.}
\begin{figure*}[t!]
    \includegraphics[width=0.85\linewidth]{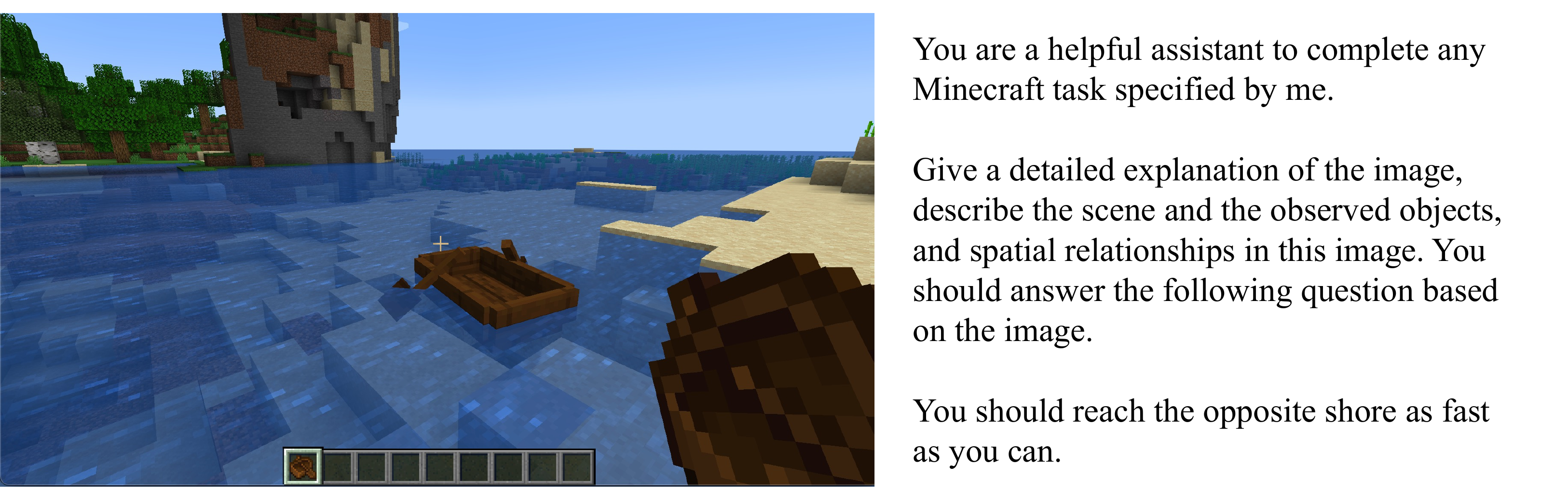}
    \caption{Embodied image caption and plan generation example task in Minecraft simulator.}
    \label{fig:appendix_minecraft}
\end{figure*}

\begin{figure*}[t!]
    \includegraphics[width=0.85\linewidth]{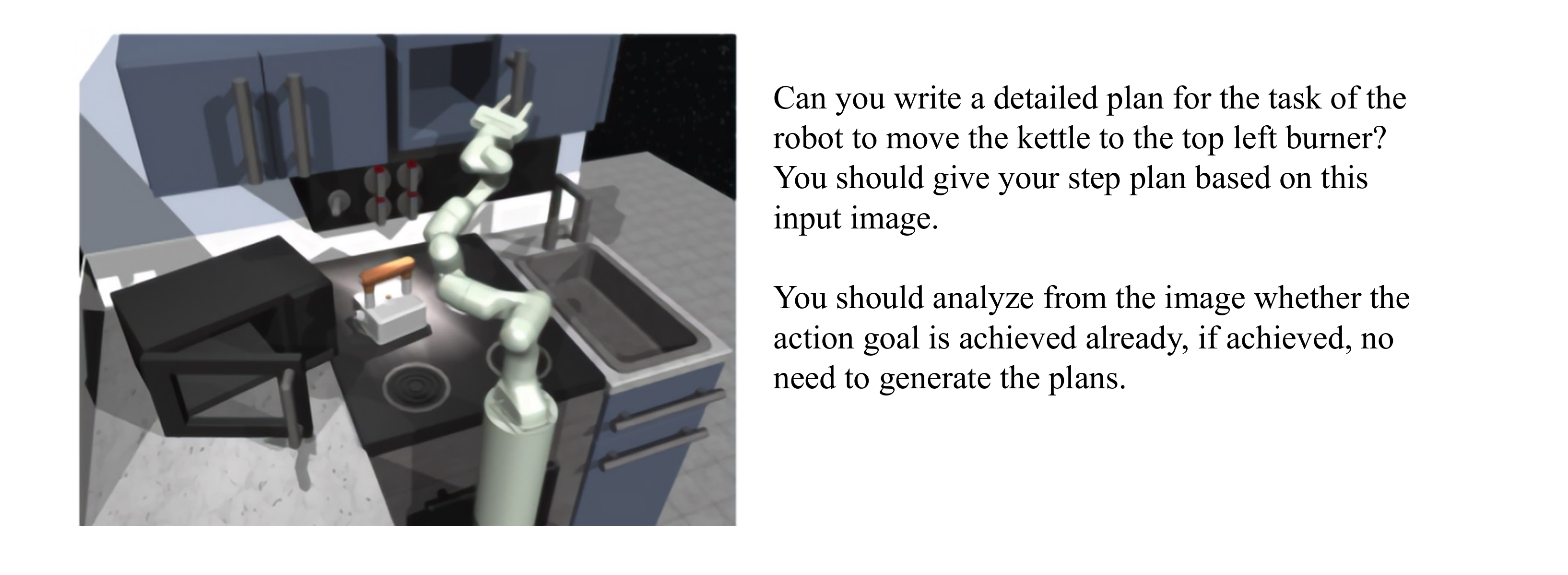}
    \vspace{-0.2in}
    \caption{Embodied image caption and plan generation example task in Franka Kitchen benchmark.}
    \label{fig:appendix_franka}
\end{figure*}

\begin{figure*}[t!]
    \includegraphics[width=0.85\linewidth]{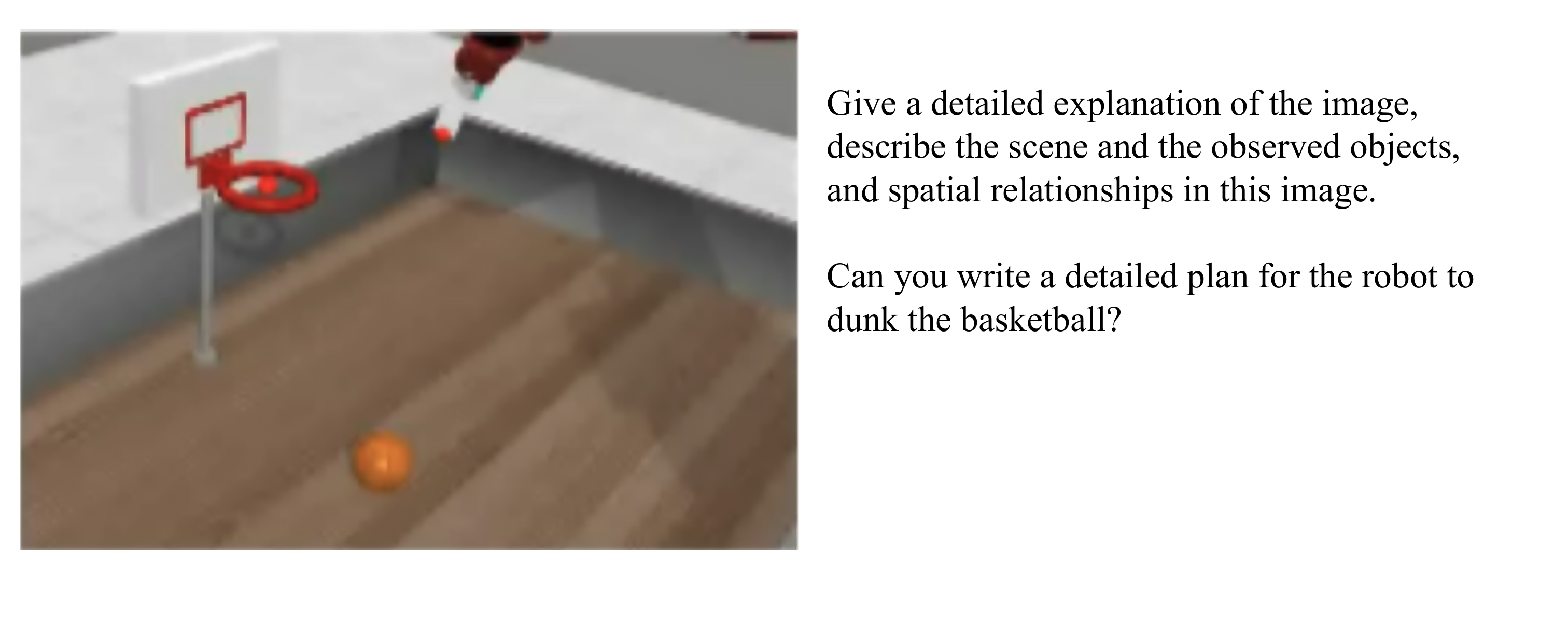}
    \caption{Embodied image caption and plan generation example task in Meta-World benchmark.}
    \label{fig:appendix_meta}
\end{figure*}

In this section, we provide quantitative evaluation results for embodied tasks in addition to the user study discussed in Section~\ref{sec:exp-embodied}. We selected some representative scenes from Minecraft, Franka Kitchen, and Meta-World benchmarks as shown in Figures \ref{fig:appendix_minecraft} through \ref{fig:appendix_meta}, and the results for these tasks are provided in Sections \ref{appendix:minecraft} to \ref{appendix:meta}.

In Figure~\ref{fig:appendix_minecraft}, the models were asked to generate feasible plans for the Minecraft agent to reach the opposite shore with a boat floating on the river. All eight models recognized the presence of the floating boat, but only LLaMA-Adapter V2, InstructBLIP, and MiniGPT-4 generated a plan that utilized the boat to help the agent reach the opposite shore more quickly.

In Figure~\ref{fig:appendix_franka}, the models needed to assist the robotic arm in moving the kettle to the top left burner, and we expected the models to analyze where the goal state was achieved from the image. Except for BLIP, all seven models provided a feasible and reasonable plan. Notably, LLaVA recognized that the goal state had already been achieved from the given image. Meanwhile, mPLUG-Owl generated some steps to deal with the situation when the goal state was achieved before execution.

For the Meta-World examples shown in Figure~\ref{fig:appendix_meta}, some models struggled with visual reasoning partially due to the low-resolution image input. mPLUG-Owl detected the basketball in the robot's hand mistakenly, while others, e.g.  LLaVA, MiniGPT-4, and Otter, mistook the robotic arm for a human.

\subsubsection{Minecraft}
\label{appendix:minecraft}
\small \begin{continueframe}

\centering
\begin{minipage}{0.98\textwidth}

{\bf BLIP:} This is a Minecraft screenshot of a boat in the water. \\

{\bf InstructBLIP:} The image depicts a Minecraft scene where the player is standing on a small island in the middle of a large body of water. The player's goal is to reach the opposite shore as quickly as possible, which involves navigating through the water and possibly encountering various obstacles along the way. In the scene, there are several objects visible, including a boat, a tree, and a rock.\textbf{ The boat appears to be partially submerged in the water, suggesting that the player may need to navigate around it or use it as a means of transportation to reach the opposite shore.} The tree serves as a landmark, providing a reference point for the player to orient themselves and determine their progress towards the opposite shore. The rock is also present, adding to the visual interest of the scene. The spatial relationship between the player, the boat, the tree, and the rock can be inferred from the image. The player is located on the small island in the middle of the water, while the boat is partially submerged nearby. The tree serves as a reference point, indicating the player's proximity to the opposite shore. The rock adds to the visual appeal of the scene, contributing to the overall sense of.  \\

{\bf LLaMA-Adapter V2:} The image shows a Minecraft scene with a small wooden boat floating on a body of water. The boat is positioned near the shore, and there is a path leading to the opposite shore. The boat appears to be in a relatively calm body of water, and it seems to be the primary mode of transportation for the player.\textbf{ The player's goal is to reach the opposite shore as fast as possible, so they will likely need to use the boat to navigate the water and reach their destination.}\\

{\bf LLaVA: } The image shows a Minecraft scene with a small boat floating in a body of water, surrounded by a sandy beach. There is also a small island in the background. The boat is positioned near the shore, and the player is standing on the beach, looking at the boat. The player's objective is to reach the opposite shore as fast as possible. The scene appears to be a part of a game, and the player's actions are likely to be controlled by the game's mechanics.\\

{\bf MiniGPT-4: } The image shows a small wooden boat floating in the water. The boat has a small cabin at the front and a mast with a sail at the rear. There is a rocky shore on one side of the boat and a sandy beach on the other. The water is calm and there are no other objects or people in the scene. The sun is shining from the top right corner of the image, casting a bright light on the water and the boat. The boat is positioned in the middle of the water, with the mast facing towards the shore. \textbf{The goal is to reach the opposite shore as fast as possible, so the player would need to navigate the boat through the water, possibly using a paddle or oars, and avoid any obstacles or hazards in the way.} \\ 

{\bf mPLUG-Owl: } In the image, a person, likely the user, is standing in the shallows near the water, looking for a way to cross over to a boat that is on the other side of the water. They appear to be in a desert biome, as evidenced by the surrounding landscape. The boat is located at a distance from the shore, and it seems to be partially underwater, suggesting that it could be a partially submerged boat. In addition to the person standing near the water, there are several rocks and trees scattered across the scene. One of the rocks is located closer to the shore, while the others are scattered farther away from it. The trees are also dispersed throughout, some closer to the shore and others further away. The presence of these objects and their spatial relationships suggest that the scene might be part of a desert-themed map or world in Minecraft. \\

{\bf Otter: } In this image, a person is standing on a raft in the middle of a body of water, with a small boat nearby on the shore. The scene is set in a game like Minecraft, where the player is trying to reach the other side of the water. The raft and the small boat are the primary objects that the player needs to utilize to cross the water safely.\textbf{ The person on the raft is standing near the boat, likely preparing to use it }to continue their journey to the other shore. \\

{\bf VPGTrans: } The image shows a small wooden boat floating in the middle of a large body of water. There is a small island on the other side of the water, and the boat is headed towards it. The water is calm, and there are no visible obstacles in the way of the boat. The boat appears to be in good condition, and there are no visible signs of damage or wear and tear. The island appears to be relatively close, and it should be possible to reach it in a short amount of time. 

\end{minipage}
\end{continueframe}

\subsubsection{Franka Kitchen}
\label{appendix:franka}
\vspace{-0.3in}
\small \begin{continueframe}
\centering
\begin{minipage}{0.98\textwidth}
{\bf BLIP:}  Move the kettle to the top left burner step. \\

{\bf InstructBLIP:} Here's a step plan: 1. Analyze the image to determine whether the action goal of moving the kettle to the top left burner has already been achieved or not. 2. If the action goal is not achieved, generate a plan for the robot to move the kettle to the top left burner. 3. Determine the initial position of the kettle on the stove. 4. Calculate the distance between the current position of the kettle and the desired position on the top left burner. 5. Based on the calculated distance, determine the necessary steps for the robot to move the kettle from its current position to the desired position on the top left burner. 6. Implement the necessary steps using the robot's movement capabilities. 7. Verify that the kettle has been successfully moved to the top left burner.  \\

{\bf LLaMA-Adapter V2:} To move the kettle to the top left burner, the robot would need to follow these steps: 1. Detect the kettle's position: The robot should first identify the kettle's current location on the kitchen counter. This can be done by analyzing the image and recognizing the kettle's shape, size, and color.2. Plan the path: The robot should then calculate the most efficient path to move the kettle to the top left burner. This could involve moving the kettle to a specific location on the counter, then rotating or tilting the kettle to align it with the burner.3. Execute the plan: The robot should then execute the plan. \\

{\bf LLaVA: } To move the kettle to the top left burner, the robot should follow these steps: 1. Identify the kettle's current position on the stove. 2. Determine the appropriate path to reach the top left burner. 3. Approach the kettle from the side, ensuring it doesn't obstruct the path. Move the kettle towards the top left burner, making sure it doesn't touch any other burners or appliances. 5. Once the kettle is positioned on the top left burner, confirm that it is stable and secure. 6. If the kettle is not yet on the top left burner, repeat steps 2-5 until the kettle is successfully placed on the burner. \textbf{In the image, the kettle is already on the top left burner, so there is no need to generate the plans. The robot has successfully moved the kettle to the desired location.} \\

{\bf MiniGPT-4: } The robot is standing in the middle of a kitchen, facing a counter with two burners on top and a sink on the left. A kettle is on the counter in front of the right burner. The robot is holding a ladle in its right arm and is reaching towards the kettle with it. Here's a plan to move the kettle to the top left burner: 1. The robot will take the ladle from its right arm and place it on the counter next to the kettle. 2. The robot will use its left arm to pick up the kettle and place it on the top left burner. 3. The robot will then take the ladle and use it to pour hot water into the kettle. Overall, the action goal is achieved in three steps.  \\ 

{\bf mPLUG-Owl: } 1. Obtain the position of the kettle on the countertop from the image. If the position of the kettle is not known, move the robot arm to scan the entire countertop area to find it. 3. After locating the kettle, the robot arm should grasp it using its gripper. 4. Using sensors such as cameras, the robot arm should navigate around other objects on the countertop and reach the desired location, which in this case is the left burner of a kitchen stove. Once the robot has arrived at the desired location, it should securely place the kettle on the top left burner. 6. The robot may need to adjust its grip on the kettle or the stove if necessary to maintain stability. 7. After the kettle is securely placed, the robot arm should release its grip and return to its original position or move back to the countertop to perform any additional tasks. Note: If it is already known that the kettle is on the top left burner, the robot may skip to step 4 or proceed to step 6 directly, based on its predefined task. \\

{\bf Otter: } In the image, the robot is already in the process of lifting the teapot with its arm and is positioning it on top of a burner on the stove. The plan would likely consist of the following steps: 1. Position the arm with the pot securely and safely on the burner. 2. Ensure that the pot is placed at the correct height and distance from the flame to avoid any accidents or damage to the pot. 3. Utilize the robot's vision system or other sensors to accurately position the pot at the desired location. 4. Once positioned, the arm should maintain a firm grip on the pot to prevent it from falling or shifting during the cooking process. 5. If the pot needs to be lowered or moved to a different burner, the same process would be repeated, with the robot carefully adjusting the position of the pot and maintaining a secure grip. The generated plan should be clear, step-by-step, and easy to follow, ensuring that the robot successfully places the pot on the desired burner while maintaining safety and efficiency. \\

{\bf VPGTrans: } To move the kettle to the top left burner, the robot should first identify the location of the kettle and the location of the top left burner. Then, the robot should use its arm to pick up the kettle and move it to the top left burner. Once the kettle is in place, the robot should use its arm to adjust the position of the kettle so that it is centered on the burner. Finally, the robot should turn on the burner to heat the water in the kettle.
\end{minipage}
\end{continueframe}
\vspace{-0.05cm}
\vspace{-0.2cm}
\subsubsection{Meta World}
\label{appendix:meta}
\vspace{-0.05cm}

\small \begin{continueframe}
\centering
\begin{minipage}{0.98\textwidth}
{\bf BLIP:}  A robot is dunking a basketball in a pool.  \\
\vspace{-0.05cm}

{\bf InstructBLIP:} The image depicts a 3D basketball game where a robot is attempting to dunk the ball. The scene features a basketball hoop, a basketball, and a robot. In order to successfully dunk the basketball, the robot needs to perform several actions. Here is a detailed plan for the robot to dunk the basketball: 1. Position the robot near the basketball hoop: The robot should be positioned close to the basketball hoop so that it can easily reach the ball when it falls through the hoop. 2. Wait for the ball to fall through the hoop: Once the ball falls through the hoop, the robot should wait for it to land on the ground. 3. Pick up the ball: The robot should pick up the ball as soon as it lands on the ground. 4. Dribble the ball towards the hoop: The robot should dribble the ball towards the hoop, aiming for the basket. 5. Shoot the ball into the hoop: The robot should shoot the ball into the hoop, attempting to dunk it. 6. Celebrate the successful dunk: If the robot successfully dunks the ball, it should celebrate its achievement. \\
\vspace{-0.05cm}

{\bf LLaMA-Adapter V2:} The image shows a basketball hoop with a basketball inside it, and a robot is positioned near the hoop. The robot is likely programmed to dunk the basketball, so it needs to execute a series of steps to achieve this goal. Here is a possible plan for the robot to dunk the basketball: 1. The robot should first identify the basketball and determine its position relative to the hoop. 2. The robot should then move towards the basketball, adjusting its speed and direction to ensure it reaches the ball at the right time. 3. As the robot approaches the basketball, it should extend its arm and grab the ball with its hand. 4. The robot should then lift its arm and place the basketball into the hoop, aiming for the basket. 5. Once the basketball is inside the hoop, the robot should release it. \\
\vspace{-0.05cm}

{\bf LLaVA: } The image shows a basketball court with a basketball hoop, a basketball, and a person holding a remote control. The scene appears to be a simulation or a video game, as the person is controlling the basketball with the remote. The basketball is positioned near the hoop, and the person is likely trying to dunk it. The remote control allows the person to simulate the dunking motion, which is not possible in real life.  \\
\vspace{-0.05cm}

{\bf MiniGPT-4: } In this image, there is a basketball player standing on a wooden floor in front of a hoop. The player is holding a basketball in one hand and is preparing to dunk it through the hoop. The player is wearing a basketball uniform, and the hoop is hanging from the ceiling. There is a small orange ball lying on the ground next to the hoop, and a second ball is sitting on the top of the hoop. To write a plan for the robot to dunk the basketball, you could use the following steps: 1. Program the robot to move to the center of the room and stand in front of the hoop. 2. Program the robot to pick up the ball on the ground with its arm. 3. Program the robot to hold the ball with one hand while it uses its other hand to dunk the ball through the hoop. 4. Program the robot to release the ball after it has gone through the hoop. 5. Program the robot to move away from the hoop after it has finished dunking the ball. Overall, this plan would involve the robot using its arm to pick up the ball on the ground, its other hand to dunk the ball through the hoop, and then releasing the ball after it has gone through.   \\ 
\vspace{-0.05cm}

{\bf mPLUG-Owl: }  The image features a robot standing on a wooden floor and holding a basketball in its hand. It seems to be attempting to perform a dunk on a basketball hoop. To achieve this, the robot needs to first position itself properly and aim at the hoop. First, it should move towards the basketball hoop by walking on its wheels. It should maintain a straight path and avoid colliding with the walls or any obstacles in its way. Once it reaches the hoop, the robot should adjust its position and align its body with the rim. It should also maintain a stable posture while maintaining the basketball in its hand. Next, the robot should jump up and extend its arm and hand towards the hoop, while simultaneously bending its knees. This will generate enough momentum to launch the basketball toward the rim. As the basketball is in its grasp, the robot should release it with a quick flick of its wrist. The ball's trajectory should follow the path of the robot's arm, and it should enter the basket with a satisfying dunk. \\
\vspace{-0.05cm}

{\bf Otter: } The image shows a small toy basketball on a basketball court, with a large toy head of a man, which appears to be a statue or a decoration, nearby. The scene suggests that the toy robot is designed to perform a dunk on the basketball using a ramp. To accomplish this, the robot would need to follow a plan that involves moving to the appropriate position on the court, using the ramp to gain elevation, and timing its jump to make contact with the basketball at the right moment. The plan would likely involve adjusting the robot's speed, angle, and trajectory to achieve a successful dunk while maintaining balance and coordination throughout the process. The specific details of the plan would vary depending on the design and programming of the robot, but the general idea would be to mimic the motion of a human dunking the ball. \\
\vspace{-0.05cm}

{\bf VPGTrans: } The image shows a basketball hoop in the middle of a basketball court. The hoop is made of metal and has a net attached to it. There is a basketball on the ground next to the hoop. The hoop is at a height of about 10 feet off the ground. The basketball is at a height of about 5 feet off the ground. There is a robot on the left side of the hoop. The robot is facing the hoop and has a net attached to its arm. The robot is at a height of about 3 feet off the ground. The robot's arm is extended and ready to shoot the basketball into the hoop.

\end{minipage}
\end{continueframe}

\section{Summary of Evaluation Datasets}\label{appendix:dataset}

\scriptsize
\begin{xltabular}{\textwidth}{>{\hsize=0.4\hsize}l|>{\hsize=1.6\hsize}X|>{\hsize=0.4\hsize}X}
\caption{Description of datasets used in our LVLM-eHub.} \label{tab:long} \\

\toprule \multicolumn{1}{c}{\textbf{Dataset Name}} & \multicolumn{1}{|c|}{\textbf{Dataset Description}} & \multicolumn{1}{c}{\textbf{Evaluation Data}} \\ \midrule 
\endfirsthead

\multicolumn{3}{c}%
{\tablename\ \thetable{} -- continued from previous page} \\
\toprule \multicolumn{1}{c}{\textbf{Dataset Name}}  & \multicolumn{1}{|c|}{\textbf{Dataset Description}} & \multicolumn{1}{c}{\textbf{Evaluation Data}} \\ \midrule  
\endhead

\bottomrule
\endfoot

\bottomrule
\endlastfoot

\multicolumn{3}{c}{Visual Perception Datasets} \\ \midrule
ImageNet1K & The ImageNet1K dataset~\cite{ImaegNet1K} consists of 1K object classes and contains 1,281,167 training images, 50 images per class for validation, and 100 images per class for testing. & 50K (val.) \\\midrule
CIFAR10 & The CIFAR10 dataset~\cite{Krizhevsky2009LearningML} has 10 classes and 6000 images per class with 5000 for training and 1000 for testing. & 10K (test) \\\midrule
Pets37 & The Oxford-IIIT Pet dataset~\cite{parkhi12a} comprises 37 categories (\textit{Pets37} for short) with 25 dog breeds and 12 cat ones and \~200 images per class. There are 7349 images in total, 3680 trainval images, and 3669 test images. &3669 (test) \\\midrule
Flowers102 & The Oxford 102 Flower dataset~\cite{Nilsback08} includes 120 flower categories (\textit{Flowers102} for short) with 40 to 258 images for each class and 8189 images in total, namely 10 images per class for both train and val and the rest for a test. & 6149 (test) \\\midrule
COCO-OC & We ask the model to count the number of a certain object appearing in the image and attend to individual objects, which is decoupled from high-level semantics and thus a more appropriate test bed for fine-grained visual understanding evaluation. We construct the dataset of this problem with images from the validation set of MSCOCO & 10000 (val) \\\midrule
COCO-MCI & We ask the model if a certain object exists in the image and attend to individual objects, which is decoupled from high-level semantics and thus a more appropriate test bed for fine-grained visual understanding evaluation. We construct the dataset of this problem with images from the validation set of MSCOCO & 10000 (val) \\\midrule
VCR-OC & Same as COCO-OC, but using images from the validation set of the VCR dataset & 10000 (val) \\\midrule
VCR-MCI & Same as COCO-MCI, but using images from the validation set of the VCR dataset & 10000 (val) \\\midrule
\multicolumn{3}{c}{Visual Knowledge Acquisition Datasets} \\ \midrule
IIIT5K & The IIIT5K~\cite{ittt5k} is an ocr dataset that contains words from street scenes and originally-digital images. It is split into 2k/3k for train/test set. & 3k (test) \\\midrule
IC13 & The ICDAR 2013 dataset~\cite{ic13} consists of 229 training images and 233 testing images, with word-level annotations provided. Specifically, it contains 848 and 1095 cropped text instance images for the train and test sets respectively. & 848 (train) \\\midrule
IC15 & The ICDAR 2015 dataset~\cite{ic15} contains 1500 images: 1000 for training and 500 for testing. Its train/test set contains 4468/2077 cropped text instance images. & 2077 (test) \\\midrule
Total-Text & The total-test dataset~\cite{total-text} contains 1555 images: 1255 for training and 300 for testing. It contains 2551 cropped text instance images in the test set. & 2551 (test) \\\midrule
CUTE80 & The CUTE80 dataset~\cite{cute80} contains 288 cropped text instance images getting from 80 high-resolution images. & 288 (all) \\\midrule
SVT & The Street View Text (SVT) dataset~\cite{svt} was harvested from google street view. It contains 350 images in total and 647 cropped text instance images for testing. & 647 (test) \\\midrule
SVTP & The SVTP dataset~\cite{svtp} contains 645 cropped text instance images. It is specifically designed to evaluate perspective-distorted text recognition. No train/test split was provided. & 645 (all) \\ \midrule
COCO-Text & The COCO-Text dataset~\cite{coco-text} we use is based on the v1.4 annotations, which contains 9896/42618 annotated words in val/train set. & 9896 (val) \\ \midrule
WordArt & The WordArt dataset~\cite{wordart} consists of 6316 artistic text images with 4805 training images and 1511 testing images. & 1511 (test) \\ \midrule
CTW & The SUCT-CTW1500 (CTW) dataset includes over 10,000 text annotations in 1500 images (1000 for training and 500 for testing) used in curved text detection. In our evaluation, we use 1572 rectangle-cropped images getting from the testing set. & 1572 (test) \\ \midrule
HOST & The heavily occluded scene text (HOST) in Occlusion Scene Text (OST) dataset~\cite{ost}. & 2416 (HOST) \\ \midrule
WOST & The weakly occluded scene text (WOST) in the OST dataset. & 2416 (WOST) \\ \midrule
SROIE & The SROIE dataset~\cite{sroie} contains 1000 complete scanned receipt images for OCR and KIE tasks. The dataset is split into 600/400 for the trainval/test set. In the KIE task, it is required to extract company, data, address, and total expenditure information from the receipt and there are 347 annotated receipts in the test set. & 347 (test) \\ \midrule
FUNSD & The FUNSD dataset~\cite{funsd} contains 199 real, fully annotated, scanned forms for the KIE task. It is split 50/149 for the test/train set. & 50 (test) \\ \midrule
NoCaps & The NoCaps dataset contains 15100 images with 166100 human-written captions for novel object image captioning. & 4500 (val) \\ \midrule
Flickr-30k & The Flickr30k dataset consists of 31K images collected from Flickr, each image has five ground truth captions. We use the test split which contains 1K images. & 1K (test) \\ \midrule
WHOOPS & The WHOOPS dataset~\cite{whoops_dataset} includes 500 synthetic and compositional images and 5 captions per image. & 2500 \\ \midrule
\multicolumn{3}{c}{Visual Reasoning Datasets} \\ \midrule
DocVQA & DocVQA~\cite{docvqa} contains 12K images and 50K manually annotated questions and answers. & 5349 (val) \\ \midrule
TextVQA & Notably, we use the latest v0.5.1 version of TextVQA~\cite{textvqa} dataset. It contains 34602 questions based on 21953 images from OpenImages' training set. Its validation set contains 5000 questions based on 3166 images. & 5000 (val) \\ \midrule
STVQA & Scene Text Visual Question Answering (STVQA)~\cite{stvqa} consists of 31,000+ questions across 23,000+ images collected from various public datasets. It contains 26074 questions in the train set and we sample 4000 samples from the train set in default order with seed 0. & 4000 (train) \\ \midrule
OCR-VQA & OCRVQA~\cite{ocr-vqa} contains 100037 question-answer pairs spanning 207572 book cover images. & 100037 (all) \\ \midrule
OKVQA & OKVQA~\cite{okvqa} is a dataset about outside knowledge visual question answering. It contains 14055 open-ended question-answer pairs in total. & 5046 (val) \\ \midrule
GQA & GQA~\cite{gqa} is a visual question-answering dataset with real images from the Visual Genome dataset. & 12578 (testdev) \\ \midrule
Visdial & Visual Dialog (Visdial)~\cite{visdial} contain images sampled from COCO2014 and each dialog has 10 rounds. In our evaluation, we treat it as a VQA dataset by splitting each dialog sample into question-answer pairs by rounds. As there are 2064 dialog samples in the validation set, we have 20640 question-answer pairs collected from the validation set. & 20640 (val) \\ \midrule
IconQA & IconQA dataset~\cite{iconqa} provide diverse visual question-answering samples and we use the test set in its multi-text-choice task. & 6316 (test) \\ \midrule
VSR & Visual Spatial Reasoning (VSR) dataset~\cite{vsr} contains a collection of caption-image pairs with true/false labels. We treat it as a VQA dataset by asking the model to answer True or False. & 10972 (all) \\ \midrule
WHOOPS & The WHOOPS dataset~\cite{whoops_dataset} encompasses 500 synthetic and compositional images and 3662 question-answer pairs in total. Specifically there is only one answer for each question. & 3662 \\ \midrule
ScienceQA IMG & ScienceQA~\cite{scienceqa} is a multimodal benchmark containing multiple choice questions with a diverse set of science topics. In our evaluation, we only use the samples with images in the test set. & 2017 (test) \\ \midrule
VizWiz & VizWiz~\cite{vizwiz} is a VQA dataset whose answers are got by asking blind people. & 1131 (val) \\ \midrule
SNLI-VE & SNLI-VE\cite{xie2019visual} extends the text entailment (TE) task into the visual domain and asks the model whether the image is semantically entailed, neutral, or contradicted to the next hypothesis. It is a three-category classification task based on Flicker30k\cite{Young2014FromID}. & 500 (dev) \\ \midrule
\multicolumn{3}{c}{Visual Commonsense Datasets} \\ \midrule
ImageNetVC & ImageNetVC\cite{xia2023imagenetvc} is a fine-grained human-annotated dataset for zero-shot visual commonsense evaluation, containing high-quality QA pairs across diverse domains with sufficient image sources. & 10000 (rank) \\ \midrule
VCR & VCR~\cite{zellers2019recognition} is a challenging multiple-choice VQA dataset that needs commonsense knowledge to understand the visual scenes and requires multiple-steps reasoning to answer the question. & 500 (val) \\ \midrule
\multicolumn{3}{c}{Object Hallucination Datasets} \\ \midrule
COCO-Random & Following \cite{li2023evaluating}, we randomly select 500 images from the validation set of MSCOCO with more than three ground-truth objects in the annotations and construct 6 questions for each image. The probing objects in the questions that do not exist in the image are randomly sampled & 3000(val) \\ \midrule
MSCOCO-Popular& Similar to COCO-Random, we randomly select 500 images and construct 6 questions for each image. But the probing objects in the questions that do not exist in the image are selected from the top-50\% most frequent objects in MSCOCO \cite{li2023evaluating}. & 3000(val)\\ \midrule
MSCOCO-Adversarial&Similar to COCO-Random, we randomly select 500 images and construct 6 questions for each image. But the probing objects in the questions that do not exist in the image are selected from the ranked objects with their co-occurring frequency and the top-50\% most frequent objects are sampled \cite{li2023evaluating}. & 3000(val)\\ \midrule
\multicolumn{3}{c}{Embodied Intelligence Datasets} \\ \midrule
Embodied AI Tasks & Minecraft~\cite{fan2022minedojo}, VirtualHome~\cite{puig2018virtualhome}, Meta-World~\cite{gupta2019relay}, and Franka Kitchen~\cite{gupta2019relay}  & selected samples \\
\bottomrule

\end{xltabular}

\end{document}